\documentclass[10pt,twocolumn,letterpaper]{article}
\usepackage{wacv}
\usepackage{times}
\usepackage{titling}
\usepackage{epsfig}
\usepackage{graphicx}
\usepackage{amsmath}
\usepackage{bbm}
\usepackage{amssymb}
\usepackage{graphicx,times,amsmath}
\usepackage{cases}
\usepackage{multirow}
\usepackage[ruled,vlined]{algorithm2e}
\usepackage{amsfonts}
\usepackage{graphicx}
\usepackage{amssymb}
\usepackage{amsfonts}
\usepackage{amsthm}
\usepackage{subfig}
\usepackage{epstopdf}
\usepackage{makecell}
\usepackage{array}
\usepackage{xspace}
\usepackage{xparse}
\usepackage{mathtools}
\usepackage{cite}
\usepackage{makecell}
\usepackage[symbol]{footmisc}
\usepackage{rotating}
\usepackage{comment}
\usepackage{booktabs}

\makeatletter
\newcommand{\nosemic}{\renewcommand{\@endalgocfline}{\relax}}
\newcommand{\dosemic}{\renewcommand{\@endalgocfline}{\algocf@endline}}
\makeatother

\newcolumntype{?}{!{\vrule width 1pt}}

\DeclareMathOperator*{\argmin}{arg\,min}

\newtheorem{theorem}{Theorem}
\newtheorem{assumption}[theorem]{Assumption}
\newtheorem{lemma}[theorem]{Lemma}
\newtheorem{proposition}[theorem]{Proposition}
\newtheorem{definition}[theorem]{Definition}

\usepackage[colorlinks=true,citecolor={black},breaklinks=true,letterpaper=true,bookmarks=false]{hyperref}

\makeatletter
\DeclareRobustCommand\onedot{\futurelet\@let@token\@onedot}
\def\@onedot{\ifx\@let@token.\else.\null\fi\xspace}
\def\eg{\emph{e.g}\onedot} 
\def\ie{\emph{i.e}\onedot} 
 
\def\etc{\emph{etc}\onedot}

\makeatother

\wacvfinalcopy
\def\confYear{CVPR 2021}

\begin{document}
\title{On the Robustness of Multi-View Rotation Averaging 
}
\author{
\begin{tabular}{c}
     Xinyi Li\\
     Department of Computer and Information Sciences\\
     Temple University\\
     Philadelphia, USA\\ 
     \{{\tt\small xinyi.Li@temple.edu}\}
\end{tabular} \and
\begin{tabular}{c}
    Haibin Ling \\
     Department of Computer Science\\
     Stony Brook University\\
     Stony Brook, USA\\
     \{{\tt\small hling@cs.stonybrook.edu}\}
\end{tabular}
}
\date{}
\maketitle

\begin{abstract}
Rotation averaging is a synchronization process on single or multiple rotation groups, and is a fundamental problem in many computer vision tasks such as multi-view structure from motion (SfM). Specifically, rotation averaging involves the recovery of an underlying pose-graph consistency from pairwise relative camera poses. Specifically, given pairwise motion in rotation groups, especially 3-dimensional rotation groups (\eg, $\mathbb{SO}(3)$), one is interested in recovering the original signal of multiple rotations with respect to a fixed frame. In this paper, we propose a robust framework to solve multiple rotation averaging problem, especially in the cases that a significant amount of noisy measurements are present. By introducing the $\epsilon$-cycle consistency term into the solver, we enable the robust initialization scheme to be implemented into the IRLS solver. Instead of conducting the costly edge removal, we implicitly constrain the negative effect of erroneous measurements by weight reducing, such that IRLS failures caused by poor initialization can be effectively avoided. Experiment results demonstrate that our proposed approach outperforms state of the arts on various benchmarks.
\end{abstract}

\section{Introduction}
Robot navigation guided by visual information, namely, simultaneous localization and mapping (SfM), primarily involves estimating and updating the camera trajectory dynamically. {\it Pose graph optimization}, as a fundamental element in SfM, devotes to iteratively fix the erroneous calculation of the camera poses due to the noisy input and misplaced data association. Conventional pose graph optimization techniques are principally fulfilled with bundle adjustment (BA), which refines camera poses by progressively minimizing the point-camera re-projection errors. Fused by state-of-the-art nonlinear programming algorithms, \eg, Levenberg-Marquardt method~\cite{Levenberg1944}, Gauss-Newton method, \etc, the camera poses and map points are successively optimized according to the sequential input images.

Motion averaging has attracted surging research interests in the 3D vision field most recently, especially on structure-from-motion (SfM) related tasks. In contrast with BA-based approaches which mainly leverage point-camera correspondences, global motion averaging aim to recover the camera poses by solving the synchronization problem, \ie, to achieve a set of camera orientations and locations which are consistent with the pairwise measurements between them. General motion averaging pipelines involve two steps: the rotation averaging based on the epipolar geometric correlation, followed by the translation averaging where the camera orientations are the solver to the previous step and considered fixed. Translation averaging is well-known to be a convex problem and all the stationary points are thus global optimal solution. However, analogous statements on robustness of existing methods are still lacking in the rotation averaging study. 

We aim to show that given cycle consistency constraints, \ie, progressively the relative rotations on a cycle structure should end with identity, the local iterative method on multiple rotation averaging problem can be initialized in a more robust manner. It is well known that Lie groups are sensitive to perturbations, even small noise on few elements in the rotation matrix can result in completely different rotations. Due to the nature of multiple rotation averaging problem, however, the measurements are normally noisy. In viewgraph of large-scale problems, the measurement noise on some edges can propagate progressively over the entire graph, resulting in unsatisfactory solutions. Moreover, for multiple rotation averaging, there does not exist a canonical direct solver and all iterative solvers rely heavily on a reasonably well initialization, convergence of the iterative solver tends to be excessively slow or eventually fails to converge.

In this work, we address all the issues mentioned above by introducing the measurement de-noising (or measurement reweighting), which is conducted before initiating the IRLS solver. Imposing the cycle consistency is essentially conducting single rotation averaging over the cyclic sub-graph. Within each cycle, the deviation of the erroneous edge is constrained by redundant measurements and hereafter diluted by allowing a set of weights on all the edges. These edge weights will then be normalized within the scale of the whole graph and exploited as the initialization in the latter IRLS iterations. Furthermore, we exploit a novel cost function in the IRLS steps such that the penalty on the erroneous measurements changes accordingly.

To validate our approach, we conduct experiments on challenging collections of unordered internet photos of various sizes and demonstrate that our proposed scheme yields similar or higher accuracy than state-of-the-art results.

In summary, the key contributions of our paper are:
\begin{itemize}
    \item We propose a robust framework to exactly recover a consistent set of camera poses in the presence of a significant amount of noisy measurements and/or outliers in SfM problems.
    \item We show that with desired connectivity on a graph, the noisy measurements can be guaranteed to yield an error upper bound
    \item We address the issues arising for existing multiple rotation averaging approaches. In particular, we demonstrate the conditions under which multiple rotation averaging schemes may fail.
\end{itemize}

\section{Related Work}
Camera pose estimation lies in the heart of monocular SfM systems, whereas the camera orientation and translation optimization consist the camera pose refinement process. Compared with conventional BA-based approaches where re-projection errors are iteratively minimized, approaches fused by rotation averaging methods have been recently proven more efficient yielding comparable or higher accuracy which greatly benefit real-time applications with limited computational power. Rotation averaging has been first introduced into 3D vision by~\cite{govindu2001combining} where the authors exploit Lie-algebraic averaging and propose an efficient and robust solver for large-scale rotation averaging problems, and was later studied in~\cite{moakher2002means}. Outperformance with the motion averaging backbone against canonical approaches stimulates numerous SfM frameworks~\cite{cui2015, zhu2018, zhu2017parallel, cui2017hsfm, Li20Hybrid, martinec2007robust, locherprogsfm}, whereas global motion averaging is conducted to simultaneously solve all camera orientations from inter-camera relative motions. In~\cite{cui2017hsfm}, the authors develop a camera clustering algorithm and present a hybrid pipeline applying the parallel-processed local increment into global motion averaging framework. Similarly, in~\cite{zhu2018}, distributed large-scale motion averaging is addressed. In~\cite{Li20Hybrid}, a hybrid camera estimation pipeline is proposed where the dense data association introduces a single rotation averaging scheme into visual SfM.

Rotation averaging~\cite{hartley2013rotation} has shown improved robustness compared with canonical BA-based approaches in numerous aspects. For instance, proper initialization plays a vital role in equipping a sufficiently stable monocular system ~\cite{tang2017gslam}, while~\cite{carlone2015initialization} addresses the initialization problem for 3D pose graph optimization and survey 3D rotation estimation techniques, where the proposed initialization demonstrates superior noise resilience. In~\cite{chatterjee2013efficient} and~\cite{chatterjee2018robust} the process is initialized by optimizing a $l_1$ loss function to guarantee a reasonable initial estimate.  It has also been shown in 567 that estimating rotations separately and initialize the 2D pose graph with the measurements provide improved accuracy and higher robustness. In\cite{kneip2012finding, kneip2014efficient} it has been exploited that camera rotation can be computed independent of translation given specific epipolar constraints. It is well known that monocular SfM is sensitive to outliers and many robust approaches~\cite{govindu2006robustness, yang2020graduated, enqvist2011non, kahl2008multiple} have thus been designed to better handle the noisy measurements. Moreover, Lagrangian duality has been reconciled in recent literature~\cite{fredriksson2012simultaneous, briales2018certifiably, carlone2015lagrangian} to address the solution optimality. A recent paper~\cite{eriksson2018rotation} shows that certifiably global optimality is obtainable by utilizing Lagrangian duality to handle the quadratic non-convex rotation constrains~\cite{wilson2016rotations} and further derives the analytical error bound in the rotation averaging framework.  
Recent work~\cite{bustos2019} and~\cite{Li20Hybrid} attempt to rely solely on rotation averaging without BA to handle SfM tasks. In~\cite{Li20Hybrid}, the authors partition the input sequence into blocks according to the pairwise covisibility and the optimization is processed hierarchically with local BA and global single rotation averaging. While~\cite{Li20Hybrid} yields high accuracy, it is demanding to handle the latency between local and global optimization and the system may suffer time overhead progressively.

Recent work~\cite{kasten2019algebraic, kasten2019gpsfm, geifman2020averaging} tackle the multiple rotation averaging problem by exploiting rank constraints on the global fundamental matrices. While the factorization-based methods show high accuracy dealing with large-scale datasets, it is much slower and costly than local iterative solver-based approaches. Our proposed approach falls into the latter category. 
Inspired by recent work~\cite{lerman2019robust} where an algorithm is proposed to solve group synchronization under significant amounts of corruption or noise, we realize that most iterative solvers for general group synchronization relies heavily on the initialization scheme and thus tends to fail in presence of noisy measurements. Analogous to~\cite{lerman2019robust}, our work also focuses on robustifying the the rotation averaging in noisy scenarios. However,~\cite{lerman2019robust} applies message passing scheme to explicitly estimate the underlying noise levels while we propose to implicitly decrease the weights on the noisy edges by enforcing cycle consistency. Other work utilizing cycle consistency includes~\cite{zach2010disambiguating} and~\cite{shen2016graph}, where~\cite{zach2010disambiguating} proposes to detect corrupted measurements by maximizing log likelihood function and~\cite{shen2016graph} classifies the edges as uncorrupted as long as they belong to any cycle-consistent cycle. In this work, instead of detecting or removing erroneous edge explicitly, we propose to implicitly avoid the negativity brought by the noisy measurements, with enforcing the reweighted graph to be cycle consistent.

\section{Theory}
\label{sec:theory}

\subsection{Definitions and Assumptions}
\label{sec:notations}
Consider a {\it simple directed graph} $\mathcal{G}:=(\mathcal{V}, \mathcal{E})$, where $\mathcal{V}=[n]=\{1, 2, \cdots, n\}$ denotes the set of {\it vertices}, $\mathcal{E}=\{(i,j)|i\neq j, i, j\in \mathcal{V}\}$ denotes the set of directed {\it edges}. We further associate a set of labels $\{\Lambda, \Sigma\}$ with $\mathcal{G}$ such that a new tuple $\mathcal{H}:=(\mathcal{V}, \mathcal{E}, \lambda, \sigma)$ is constructed in a way that $\lambda:\mathcal{V}\rightarrow \Lambda$, $\sigma: \mathcal{E} \rightarrow \Sigma$. Specifically, as we primarily focus on the synchronization with 3-dimensional rotation groups (\ie, $\mathbb{SO}(3)$), we assume $\Lambda \subseteq \mathbb{SO}(3)$ and $\Sigma \subseteq \mathbb{SO}(3)$ unless stated otherwise for the rest of the paper. Furthermore, we assume $(j,i)\in \mathcal{E}$ if and only if $(i,j) \in \mathcal{E}$ for all $i\neq j \in \mathcal{V}$ due to the nature of the problem. We thus have $\sigma (i,j) = \sigma (j,i)^\top = \sigma (j,i)^{-1}$. For ease of notation, we henceforth let $\mathcal{E}$ be the set of {\it undirected} edges associated with ordered labeling $\sigma$. 

With the notations above, we further clarify the problem as follows. Consider an $n$-view scene with $m$ measurements of relative rigid motions between the cameras. Specifically, node $i \in \mathcal{V}$ in $\mathcal{H}$ denotes the $i^{\text th}$ view, $\lambda(i) \in \mathbb{SO}(3)$ denotes the absolute camera rotation at $i$;
$(i,j)\in \mathcal{E}$ if there exists relative transformation measurements between the $i^{\text th}$ view and the $j^{\text th}$ view, where $|\mathcal{V}|=n$, $|\mathcal{E}|=m$. Denote $\tilde{\sigma} (i,j)\in \mathbb{SO}(3)$ for the measurement of relative rotation from $i$ to $j$ (and $\tilde{\sigma} (j,i)$ for the opposite direction) and $\bar{\sigma} (i,j)$ for the corresponding ground truth. The {\it edge measurement error} is denoted by $\epsilon (i,j):=d(\tilde{\sigma}(i,j), \bar{\sigma}(i,j))$, where $d:\mathbb{SO}(3) \times \mathbb{SO}(3) \rightarrow \mathbb{R}_{0}^{+}$ denotes some metric. 
\begin{definition}[graph consistency]
\label{def:graphconsistency}
Given a labeled graph $\mathcal{H}=(\mathcal{V}, \mathcal{E}, \lambda, \hat{\sigma})$ and some metric $d$ as defined above. For a given $\lambda$, $\mathcal{H}$ is $\lambda$-consistent if and only if
\begin{equation}
    \begin{aligned}
        \lambda (i) \cdot \hat{\sigma}(i,j) = \lambda (j),\;\; \forall (i,j)\in \mathcal{E},
    \end{aligned}
\end{equation}
which is equivalent with
\begin{equation}
\label{eq:sum0}
    \sum_{(i,j)\in \mathcal{E}} d(\lambda (i)\hat{\sigma}(i,j), \lambda _j)=0.
\end{equation}
\end{definition}

The multiple rotation averaging problem studied in this paper can thus be considered as the group synchronization process, where we recover $\lambda$ by imposing graph consistency on $\mathcal{H}$. Since no direct solver for synchronization on $\mathbb{SO}(3)$ exists, it 
is conventionally formulated as the optimization problem, which can be solved iteratively. Specifically, Eq.~\ref{eq:sum0} is equivalently as
\begin{equation}
\label{eq:nocost}
    \argmin_{\lambda} \sum_{(i,j)\in \mathcal{E}} d(\lambda (i)\hat{\sigma}(i,j), \lambda _j),
\end{equation}
in practice, Eq.~\ref{eq:nocost} is rarely solved in its original form. Instead, a cost function is often applied into the optimization, equivalently as solving
\begin{equation}
    \label{eq:main}
    \mathbf{\argmin_{\lambda} \sum_{(i,j)\in \mathcal{E}} \rho (d(\hat{\sigma}(i,j), \lambda _i^{-1} \lambda _j ))},
\end{equation}
where $\rho(\cdot)$ is some cost function. In the following analysis, we use {\it geodesic distance} for $d(\cdot, \cdot)$ by convention and $l_1$ cost function for measurement correction as the low penalty makes the solver more robust than the counterparts, we use a convex cost function for IRLS as described in \S\ref{sec:irls}. Refer to \S\ref{sec:ablationnorm} for the comparison of the performances with different cost functions.

\begin{figure}
\setlength{\belowcaptionskip}{-10pt}
    \centering
    \includegraphics[width=\linewidth]{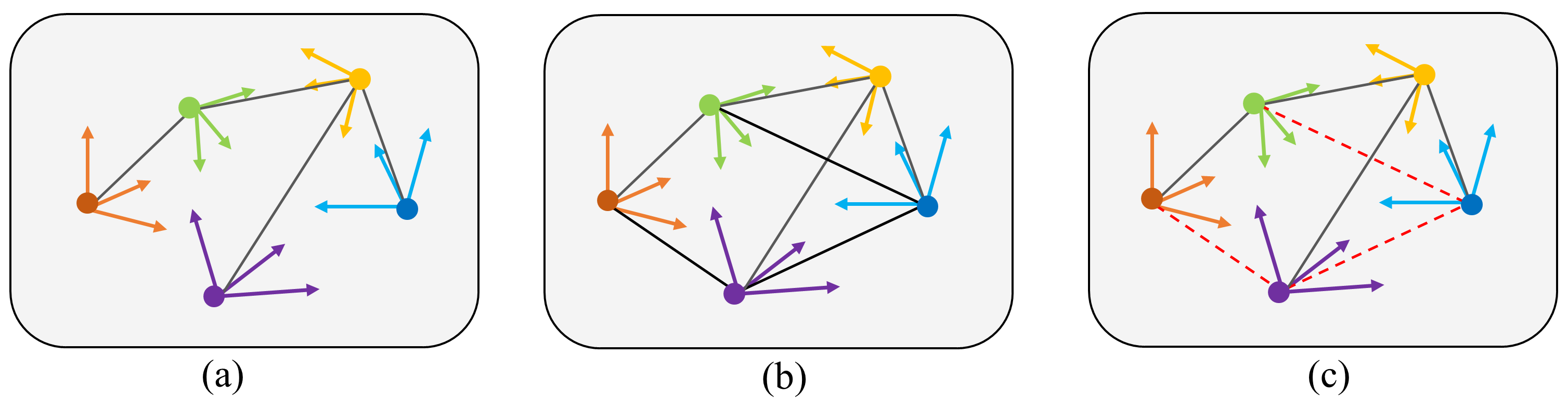}
    \caption{Common cases of group synchronization in structure from motion tasks. The coordinates represent cameras with different motion, where solid black lines represent there exists measurable relative motion between the two frames, red dashed lines represent that the measurements of relative motion contain high noise though they are measurable. In (a), the pose-graph does not contain cycles so the error compensation based on cycle consistency cannot be conducted. In (b), the pose-graph contains cycles but the measurements are noise-free. In (c), the pose-graph contains cycles but some of the edges are noisy -- in this paper, we focus on group synchronization in this case.}
    \label{fig:my_label}
\end{figure}
To solve Eq.~\ref{eq:main}, we want to first make sure that $\hat{\sigma}$ is reliable to some extent as Eq.~\ref{eq:main} is well-known to be sensitive to measurement outliers. Accordingly, We begin the analysis by introducing cycle consistency.
\begin{definition}[cycle consistency]
\label{def:cycleconsistency}
Given an edge-labeled graph $\mathcal{H}=(\mathcal{V}, \mathcal{E}, \sigma)$ as defined above, for $i\in \mathcal{V}$, denote $\mathcal{C}_i:=\{\mathcal{C}_i^1, \mathcal{C}_i^2, \cdots \}$ for the union of all the cycles containing $i$.
Then $\mathcal{H}$ is cycle consistent if and only if 
\begin{equation}
\label{eq:zeroconsistency}
\begin{aligned}
    \sigma(i,i_1)\cdot \prod_{j=1}^{m_k -1} &\sigma({i_j}, i_{j+1}) \cdot \sigma(i_{m_k},i)=\mathbb{I}_\sigma  \\
    &\forall i\in \mathcal{V}, \forall \mathcal{C}_i^k \in \mathcal{C}_i,
\end{aligned}
\end{equation}
where $m_k = |\mathcal{C}_i^k|$, $\mathbb{I}_\sigma$ denotes the identity mapping under $\sigma$.
\end{definition}

Generally, cycle consistency implicitly infers that all the edge measurements are precise in the graph, which rarely occurs in reality. Indeed, for most computer vision related tasks, input de-noising is one of the most important steps. For example, camera relative motion measurements are essentially computed with pairwise geometric constraints in common SfM pipelines. Since these constraints are normally derived from the photometric information, such as feature matching, the matching outliers can bring significant noise in the measurements and thus result in corrupted, erroneous edge labelling. Most previous work use RANSAC to detect and remove feature outliers and/or edge outliers, the process itself is, however, extremely costly and slow when the graph is sufficiently large. In this work, we propose to reduce the negative effects of noisy edges to robustify the solver, by implicitly enforcing cycle consistency, with reweighting both the vertices and the edges. 

Given that a non-trivial multiple rotation averaging problem involves at least one cycle in the pose-graph, we relax Definition~\ref{def:cycleconsistency} into the following.
\begin{definition}[$\epsilon$-cycle consistency]
\label{def:epsconsistency}
Let $\mathcal{H}$, $\mathcal{C}_i$ be defined as in Definition~\ref{def:cycleconsistency}, we say $\mathcal{H}$ is $\epsilon$-cycle consistent if 
\begin{equation}
\label{eq:epsconsistency}
    \max_{\substack{i\in \mathcal{V}\\ \mathcal{C}_i^k \in \mathcal{C}_i}} d(\sigma(i,i_1)\cdot \prod_{j=1}^{m_k -1} \sigma({i_j}, i_{j+1}) \cdot \sigma(i_{m_k},i),\mathbb{I}_\sigma)\leq \epsilon
\end{equation}
\end{definition}

\begin{figure}
    \centering
    \includegraphics[width = 0.65\linewidth]{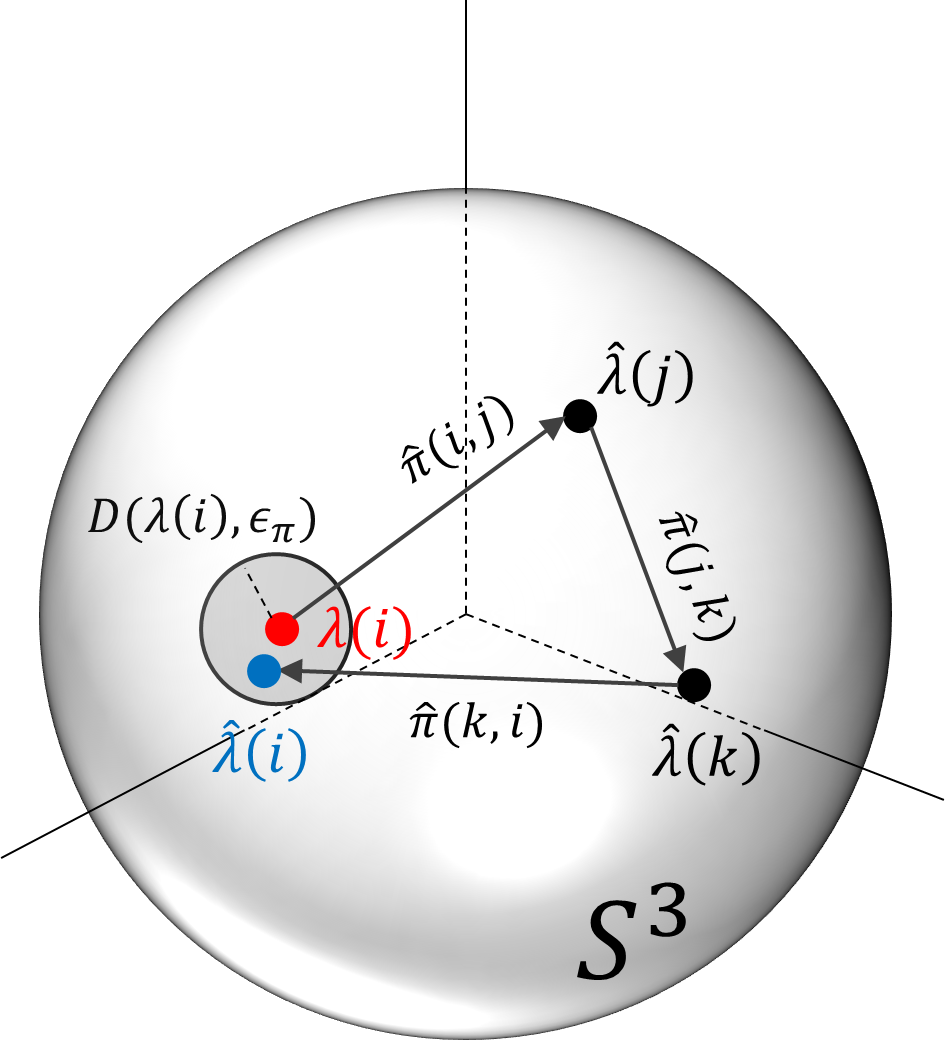}
    \caption{Illustration of $\epsilon$-cycle consistency of cycle $(i,j,k)$ with respect to $i$ on the unit 3-sphere embedded in $\mathbb{R}^4$. $\lambda (i)$ is an arbitrary rotation, $\hat{\pi}$ denotes the equivalent transformation of $\hat{\sigma}$ on the manifold. $D(\lambda (i), \epsilon _{\pi})$ denotes the disk on $\mathcal{S}^3$ centered at $\lambda(i)$ with geodesic radius $\epsilon _{\pi}$.} 
    \label{fig:s2sphere}
\end{figure}
To visualize the geometric meaning of Definition~\ref{def:epsconsistency}, let us consider a unit 3-sphere $\mathcal{S}^3$ and a simple cyclic graph with vertices $(i,j,k)$. Given the mutual relative rotation measurements $\hat{\sigma} (i,j),\hat{\sigma} (j,k),\hat{\sigma} (k,i)$, there exists a set of 3-sphere rotations $\hat{\pi} (i,j), \hat{\pi} (j,k), \hat{\pi} (k,i)$, which are isomorphic to its $\mathbb{SO}(3)$ counterparts\footnote{We are using sloppy notations here, formally, there exists a mapping $\pi:\mathcal{S}^3 \rightarrow \mathcal{S}^3$ such that $\pi \cong \mathbb{SO}(3)$}. As we only consider the relative rotations here, without loss of generality, assume an arbitrary $\lambda (i) \in \mathcal{S}^3$. In this case, $\hat{\lambda} (j)=\lambda (i)\hat{\pi}(i,j)$ is on the 3-sphere and so are the following rotations henceforth. After successive rotations, we arrive at that,
it is desired to have $\hat{\lambda}(k) \hat{\pi} (k,i) \in \mathcal{D}(\lambda (i),\epsilon) $, where $\mathcal{D}(p,r):=\{q\in \mathcal{S}^3|d(q,p)\leq r\}$. That is, the computed $\hat{\lambda} (i)$ should locate in the $\epsilon$-disk of $\lambda (i)$ on $\mathcal{S}^3$. See Fig.~\ref{fig:s2sphere} for the visualization. Moreover, now we can rewrite Eq.~\ref{eq:epsconsistency} of $(i,j,k)$ in its sphere form, that is
\begin{equation}
\label{eq:epspi}
    d_{\pi}(\hat{\pi}(i,j) + \hat{\pi}(j,k) + \hat{\pi}(j,k), \mathbb{I}_\pi ) \leq \epsilon _{\pi},
\end{equation}
where $d_{\pi}, \mathbb{I}_{\pi}, \epsilon _{\pi}$ are defined on $\mathcal{S}^3$ in analogous manner with those defined above on $\mathbb{SO}(3)$. More details on the isomorphism can be found in the supplementary materials.

The main reason of exploiting the $\mathcal{S}^3$ isomorphism is that, the proper vector field associated with the sphere enables the reweighting scheme, such that the {\it redundant error distribution} over all measurements in cycles can be efficiently propagated. The weighting can also be uniquely projected back to $\mathbb{SO}(3)$ with $(\alpha ^w, \beta ^w, \gamma ^w)$, where $(\alpha, \beta, \gamma)$ denotes the corresponding Euler angles.

\begin{assumption}[reweighted $\epsilon$-cycle consistency condition]
\label{assump:cycleconsistency}
Given $\mathcal{H}$, for all $(i,j)$ where there exists at least one cycle $\mathcal{C}\subseteq \mathcal{H}$ such that $(i,j) \in \mathcal{E}_{\mathcal{C}}$, there exists a set of weights $w_{ij}$ associated with the measurements such that the $\epsilon$-cycle consistency is satisfied for $\mathcal{H}$, for an arbitrary $\epsilon>0$. Formally,
\begin{equation}
    \label{eq:spherereweight}
    d_{\pi}(\sum_{(i,j)\in \mathcal{C}} w_{ij}\hat{\pi}(i,j),\mathbb{I}_{\pi})\leq \epsilon _{\pi},
\end{equation}
equivalently with
\begin{equation}
    \label{eq:rotationreweight}
     d( \prod_{(i,j)\in \mathcal{C}\setminus (k,i)} \hat{\sigma}^w(i,j) \cdot \hat{\sigma}^w(k,i),\mathbb{I}_\sigma)\leq \epsilon.
\end{equation}
\end{assumption}

\subsection{Theoretical Results}
\label{sec:theoreticalresults}
In the following analysis, we assume that $\mathcal{H}$ is connected though it may not hold valid in generalized group synchronization problem. In SfM tasks, isolate vertices in pose-graph rarely occur and will be discarded whence they do. Indeed, we further assume there are more than ${m}/{2}$ edges which are contained in at least one cycle, where $m=|\mathcal{E}|$. We summarize the assumptions and have the following proposition.
\begin{proposition}
\label{prop:weightexistence}
Given a connected $\mathcal{H}$ with $|\mathcal{E}|=m$, assume that there are $p(>\frac{m}{2})$ edges which are contained in at least one cycle and denote the set of the edges as $\mathcal{E}_p$. Then if Assumption~\ref{assump:cycleconsistency} holds for given $\epsilon$, there exists  $\{w_{ij}\}$ such that
\begin{equation}
    w_{ij} d(\hat{\sigma} (i,j), \lambda _i^{-1}\lambda _j ) < \epsilon, \;\; \forall (i,j) \in \mathcal{E}_p .
\end{equation}
\end{proposition}
\begin{proof}
Assume that there does not exist self-consistent corrupted cycle, then for a given edge $(i,j)$, within any cycle $\mathcal{C}\subseteq \mathcal{H}$ that contains $(i,j)$, Assumption~\ref{assump:cycleconsistency} gives that
\begin{equation}
    d( \prod_{(i,j)\in \mathcal{C}\setminus (k,i)} \hat{\sigma}^w(i,j) \cdot \hat{\sigma}^w(k,i),\mathbb{I}_\sigma)\leq \epsilon.,
\end{equation}
equivalent with
\begin{equation}
    d( \lambda (i)\prod_{(i,j)\in \mathcal{C}\setminus (k,i)} \hat{\sigma}^w(i,j) \cdot \hat{\sigma}^w(k,i),\lambda (i))\leq \epsilon.
\end{equation}
for any $\lambda(i)$ as defined.

Denote $\delta _{ij}$ for the deviation error on edge $(i,j)$, then given the error upper bound, the worst case scenario for $\delta _{ij}$ is that $\delta _{rs}=0$ for any other $(r,s)\in \mathcal{C}$, \ie , $\hat{\sigma}^w (r,s)=\hat{\sigma} (r,s)=\sigma (r,s)$, where the weights on all other edges are trivial. The inequality above can thus be rewritten into
\begin{equation}
    d(\lambda (j) \hat{\sigma}^w (j,i), \lambda (i)) =w_{ij}d( \hat{\sigma} (j,i), \lambda (i)\lambda (j)^{-1})\leq \epsilon.
\end{equation}
\end{proof}
\begin{lemma}
Let $\mathcal{H}$ be as defined in Prop.~\ref{prop:weightexistence}, then the $(m-p)$ `acyclic' edges must belong to some tree with root contained in $\mathcal{V}_p$, where $\mathcal{V}_p:=\{i\in \mathcal{H}|\exists j\in \mathcal{H} \;{\text s.t. }(i,j)\in \mathcal{E}_p\}$.
\end{lemma}

The lemma above immediately leads to the fact that, only the optimizations conducted on $\mathcal{H}_p$ is effective due to the lack of measurement redundancy in $\mathcal{H}_{n \setminus p}$, \ie, the cost function on $\mathcal{H}_{n \setminus p}$ can always be made zero (\ie trivial optimization). We summarize the observation into the following lemma.

\begin{lemma}
Let $\mathcal{H}$ and $\mathcal{E}_p$ be as defined in Prop.~\ref{prop:weightexistence}, then solving Eq.~\ref{eq:main} on $\mathcal{H}$ is equivalent with that on $\mathcal{H}_p:=(\mathcal{V}_p, \mathcal{E}_p, \lambda, \sigma)$.
\end{lemma}

Combining Prop.~\ref{prop:weightexistence} and the lemmas above, we make the following claim on the convexity of solving Eq.~\ref{eq:main}.
\begin{theorem}[local convexity]
\label{thm:localconvexity}
The multiple rotation averaging problem defined on $\mathcal{H}^w$ is equivalent with that defined on $\mathcal{H}_p^w$, and 
locally convex everywhere on the domain of some cost function $\rho$ if $p\geq c(1-\epsilon)^q n$, where $c,q>0$ depends on $\rho$,
$\mathcal{H}^w$ is the reweighted $\mathcal{H}$ according to Prop.~\ref{prop:weightexistence}, defined as $\mathcal{H}^w:=(\mathcal{V}, \mathcal{E}^w, \hat{\sigma}, \lambda)$. 
\end{theorem}
To avoid the ambiguity in notations, note that $(\mathcal{V}, \mathcal{E}^w, \hat{\sigma})$ is equivalent with $(\mathcal{V}, \mathcal{E}, \hat{\sigma}^w)$ with the weights defined in different subspaces.
It has been well proven that the (local) convexity of the rotation averaging problem depends on the graph connectedness and the noise level of the input measurements. Intuitively, it is more difficult to achieve the solver optima with a weakly-connected graph\footnote{Note that the weak-connectedness here means that not sufficiently many vertices are connected, which is not the same with that in general graph theory terms.} and a large number of erroneous measurements. Moreover, consider an edge $(i,j)$ which is present in $k$ cycles in $\mathcal{H}$, then the weighted measurement $\hat{\sigma}^w(i,j)$ essentially gets closer to the noise-free ground truth $\sigma^* (i,j)$ as $k$ increases. Detailed proofs to Thm.~\ref{thm:localconvexity} is provided in the supplementary materials.

Given the local convexity given by Thm.~\ref{thm:localconvexity}, it immediately follows that there exists a globally optimal solution to the objective function with some cost function $\rho$, which leads to the following statement.

\begin{theorem}[cost function]
Consider the optimization problem defined in Eq.~\ref{eq:nocost}, denote $\lambda ^*$ as the optimal solution set. Then there exists a convex, differentiable $\rho :\mathbb{R} \rightarrow \mathbb{R}$ such that Eq.~\ref{eq:main} converges to $\lambda _{\rho}^* \sim \lambda ^*$ with equality up to a global action.
\end{theorem}

In developing the iteration updates, $l_1$ and $l_2$ cost functions have been prevailing in the previous approaches. Among which $l_1$ shows a stronger robustness in handling the problem of noisy or corrupted nature, but it tends to take significantly long time when dealing with large-scale problems. While $l_2$ cost function shows a superior convergence speed, $l_2$-type cost functions are more sensitive to outliers as the gradient for general $l_2$ cost is unbounded.  
In our approach, we use $\rho(x) = x \exp(\tau x)$ as the cost function, where $\tau$ represents the penalty parameter. Derivation of the update rules according to conventional IRLS algorithm are given in \S~\ref{sec:irls}. 
\section{Method}
In this section we introduce the optimization scheme we propose to solve the multiple rotation averaging problem with addressing on the measurement correction. Specifically, we first depict the measurement de-noising algorithm owing to the cycle consistency constraints introduced in \S\ref{sec:measurement}, followed by the implementation of the IRLS solver in our proposed scheme, described in \S\ref{sec:irls}.   
\subsection{Measurement De-noising with Cycle Consistency Constraints}
\label{sec:measurement}
Recall Assumption~\ref{assump:cycleconsistency} where we presume that $\mathcal{H}$ is equipped with $\epsilon$-cycle consistency in solving Eq.~\ref{eq:main}. By enforcing cycle consistency, the latter solver (\eg IRLS, ADMM, \etc) benefits significantly for the following reasons. First, Lie groups are well-known to be sensitive to perturbations, \ie, even small noise on few elements in the rotation matrix can result in completely different rotations. Due to the nature of the multiple rotation averaging problem, however, the measurements are normally noisy. In viewgraph of large-scale problems, the measurement noise on some edges can propagate progressively over the whole graph, resulting in unsatisfactory solutions. 
Moreover, for multiple rotation averaging, there does not exist a canonical direct solver and all iterative solvers rely heavily on a reasonably well initialization. With severely noisy measurements and a trivial set of initialization, convergence of the iterative solver tends to be excessively slow or eventually fails to converge. 

In this work, we address all the issues mentioned above by introducing the measurement de-noising (or measurement reweighting), which is conducted before initiating the IRLS solver. Imposing the cycle consistency is essentially conducting single rotation averaging over the cyclic sub-graph. Within each cycle, the deviation of the erroneous edge is constrained by redundant measurements and hereafter diluted by allowing a set of weights on all the edges. These edge weights will then be normalized within the scale of the whole graph and exploited as the initialization in the latter IRLS iterations. 

In details, given a graph $\mathcal{H}$ with relative rotation measurements as group ratios, we first conduct cycle detection to find the set of cycles $\mathcal{C}:=\{\mathcal{C}_k|\mathcal{C}_k \subseteq \mathcal{H}\}$. As a cycle might contain a large number of edges and the computation is excessively costly, in practice, we instead randomly pick three vertices for $\lfloor \sqrt{|\mathcal{V}_{\mathcal{C}_k}|} \rfloor$ times. Assume, for example, $i,j,k \in \mathcal{V}_{\mathcal{C}_k}$ are selected in an iteration, then $\hat{\sigma}(i,j) = \prod _{(r,s)\in path_i^j} \hat{\sigma} (r,s)$, where path$_i^j$ denotes the connected edge from $i$ to $j$ and we denote $\hat{\sigma} (j,k)$ and $\hat{\sigma} (k,i)$ in the same manner. With the weights computed as depicted in 
\S\ref{sec:notations}, we end up with `triangle consistency' with the current vertices. These weights are then propagated along the corresponding path with scale normalization. In the experiments we simply take the path weight average over the degree of the path and achieve a satisfactory weight initialization. In the weight solution, we use $l_1$ norm for the residual computation to further increase the robustness of our proposed scheme. In contrast with edge removal schemes, we implicitly penalize the erroneous measurements by imposing smaller weights on them, as to avoid the high computational costs in edge noise removal.

It is common for an edge to appear in multiple cycles, in this case, the final edge weight is calculated with the weighted mean with respect to the according cycle sizes, \ie , consider an edge $(i,j)$ such that there is a weight set $\{w_{ij}^k\}$ for $(i,j)$ according to cycle consistency in different $\mathcal{C}_k$'s, then the final weight for $(i,j)$ is
\begin{equation}
    w_{ij} = \sum _k |\mathcal{E}_{\mathcal{C}_k}|w_{ij}^k / \sum _k |\mathcal{E}_{\mathcal{C}_k}|.
\end{equation}
\subsection{IRLS Optimization}
\label{sec:irls}
To solve Eq.~\ref{eq:main}, we exploit the conventional IRLS solver with the cost function $\rho(x) = x \exp (\tau x)$. As we mentioned before, it is well-known that the solution accuracy and convergence speed both rely heavily on the initialization. Since we already have a set of reasonable weights owing to enforcing cycle consistency, our solver is significantly more robust than previous work solely with RANSAC filtration. Comparisons with the RANSAC-based initialization schemes are provided in \S\ref{sec:ablationconsistency}.  

Now we briefly describe the update rule we employ in constructing the IRLS solver, the full derivation is provided in the supplementary materials. The algorithm is provided in Alg.~\ref{alg:MRAIRLS}.

\begin{algorithm}[h!]
\SetAlgoLined
\textbf{Input:} Set of relative transformations measurements $\{\hat{\sigma} (i,j)\}$, threshold $\alpha$\;
\textbf{Output:} Set of absolute rotation $\{\lambda _i\}$\; 
\textbf{Initialization:} Set residual $res =10^8$, iteration number $k=1$, $\tau = 1$, $\{\lambda _i\}$ = identity matrix, $w_{ij}$ from cycle consistency step\;
 \While{$res>\alpha$}{
 1. $k = k+1$\;
 2. $\delta _{ij} = \delta _j^{-1} \lambda _j ^{-1} \sigma (i,j) \lambda _i \delta _i$\;
 3. $r_{ij}\leftarrow \delta _{ij} \exp (\tau \delta _{ij})$\;
 4. $\phi _{ij} \leftarrow (1+r_{ij})\exp (\tau r_{ij})$\;
 5. $h_{ij} \leftarrow (1+\tau ^2 r_{ij}+\tau) \exp (\tau r_{ij})$\;
 6. $s \leftarrow \phi ^\top \phi /\|\phi ^\top h \phi\|$\;
 7. $w_{ij} \leftarrow s\phi _{ij}$\;
 8. $\lambda _i \leftarrow \sum w_{ij} \frac{\log (\lambda _i^{-1} \lambda _j)}{\|\log (\lambda _i ^{-1} \lambda _j)\|}$\;
 9. $res = \sum w_{ij} \delta _{ij} \exp (\tau w_{ij} \delta _{ij})$\;
 10. $\tau = 1/k$\;
 }
 \caption{MRA-Robust IRLS}
 \label{alg:MRAIRLS}
\end{algorithm}

Denote $\|\cdot \|$ as the equivalent angle for $d(\cdot, \cdot)$. Recall our objective function Eq.~\ref{eq:main}, assume that we update $\lambda$ with $\delta \lambda$, \ie, for $\lambda _i$, the updated $\lambda _i$ is $\lambda _i \delta _i$, where $\delta _i$ denotes the update. Then at one iteration, it is equivalent to solve Eq.~\ref{eq:main} as to minimize the following 
\begin{equation}
\sum_{(i,j)\in \mathcal{E}_p} \rho (\|\delta _j^{-1} \lambda _j ^{-1} \sigma (i,j) \lambda _i \delta _i\|).
\end{equation}
Then for an edge $(i,j)$, the residual $r_{ij}$ is thus
\begin{equation}
    r_{ij} = \rho(\|\delta _j^{-1} \lambda _j ^{-1} \sigma (i,j) \lambda _i \delta _i\|) = \rho(\delta_{ij}).
\end{equation}
Denote $\phi _{ij}$ as the gradient, $h_{ij}$ as the hessian of $\rho$ , the step size $s$ is computed as
\begin{equation}
    s = \|\phi\|^2 / \|\phi ^\top h  \phi \|.
\end{equation}
The updated weight $w_{ij}$ and $\lambda _i$ are then
\begin{align}
    w_{ij} &= s \phi _{ij},\\
    \lambda _i = \sum_{(i,j)\in \mathcal{E}_p} &w_{ij} \dfrac{\log (\lambda _i^{-1} \lambda _j)}{\|\log (\lambda _i^{-1} \lambda _j)\|}.
\end{align} 
In practice, we observe that with $\tau _k = 1/k$ where $k$ denotes the iteration number, the convergence displays a quadratic convergence and changes to linear at the end. The behavior is expected from the construction of $\rho (x)$, as we desire the penalty to be high at the beginning and subtle by the end of the iterations. 
\section{Experimental Results}
{\it System Configuration} All of our experiments are conducted on a PC with Intel(R) i7-7700 3.6GHz processors, 8 threads and 64GB memory. The bundle adjustment is conducted by applying Ceres library~\cite{ceres-solver}. 

{\it Methods and Datasets} We compare our proposed approach
with recent state-of-the-art approaches including~\cite{chatterjee2018robust, chatterjee2013efficient, lerman2019robust}. The approaches are tested on the Photo Tourism Dataset~\cite{wilson_eccv2014_1dsfm} and the KITTI Odometry~\cite{Geiger2012CVPR}. In our experiments, all the optimization steps are modified from public libraries~\cite{ceres-solver, kummerle2011g, l1magic} in C++.
\subsection{Quantitative Results}
\begin{table}[th!]
	\begin{center}
	\caption{Experiment results on Tourism Dataset~\cite{wilson_eccv2014_1dsfm}. In the table, $\Delta \bar{\text {deg}}$ and $\Delta \hat{\text {deg}}$ denote the mean and median error in degrees, respectively; runtime is in seconds and number of iterations denote the iterations to initialize + iterations for the calculation. Full result in supplementary materials.}
	\label{table:result}
	\addtolength{\parskip}{-0.5cm}
	\resizebox{\linewidth}{!}{%
	\begin{tabular}{ccccccc}
		\toprule[0.5mm]
         & &\textbf{\makecell{IRLS\\ \cite{chatterjee2013efficient}}} & \textbf{\makecell{Robust-IRLS\\ \cite{chatterjee2018robust}}} & \textbf{\makecell{MPLS\\ \cite{lerman2019robust}}} &
         \textbf{\makecell{Ours-$l_2$}} &
         \textbf{\makecell{Ours\\}} \\ \midrule[0.3mm]
          \multirow{4}{*}{\rotatebox{90}{Alamo}} &$\Delta \bar{\text {deg}}$ & 3.64 & 3.67 & 3.44 & 3.54 & \textbf{3.39}\\
          & $\Delta \hat{\text {deg}}$ &1.30 & 1.32 & \textbf{1.16} & 1.29 & \textbf{1.16}\\
          & runtime & \textbf{14.2} & 15.1 & 20.6 & 16.3 & 14.7\\
          & iteration &10+8 & 10+9 & 6+8 & 4+8 & \textbf{4+6}\\          \midrule[0.3mm]
          \multirow{4}{*}{\rotatebox{90}{Ell. Is.}} &$\Delta \bar{\text {deg}}$ & 3.04 & 2.71 & 2.61 & 2.39 & \textbf{2.25}\\
          & $\Delta \hat{\text {deg}}$ &1.06 & 0.93 & 0.88 & 0.82 & \textbf{0.64}\\
          & runtime & 3.2 & 2.8 & 4.0 &2.7& \textbf{2.3}\\
          & iteration &10+9 & 10+13 & 6+11 & 6+12 & \textbf{5+7}\\\midrule[0.3mm]
          \multirow{4}{*}{\rotatebox{90}{Mont. N.D}} &$\Delta \bar{\text {deg}}$ & 1.25 & 1.22 & \textbf{1.04} & 1.12 & \textbf{1.04}\\
          & $\Delta \hat{\text {deg}}$ &0.58 & 0.57 & 0.51 & 0.51 & \textbf{0.47}\\
          & runtime & \textbf{6.5} & 7.3 & 9.3 & 7.6 & 7.1\\
          & iteration &10+9 & 10+13 & 6+11 & 6+12 & \textbf{5+7}\\ \midrule[0.3mm]
          \multirow{4}{*}{\rotatebox{90}{Not. Da.}} &$\Delta \bar{\text {deg}}$ & 2.63 & 2.26 & 2.06 & 2.19 & \textbf{1.98}\\
          & $\Delta \hat{\text {deg}}$ &0.78&0.71&0.67&0.71 & \textbf{0.62}\\
          & runtime & \textbf{17.2}&22.5 & 31.5 & 24.5 & 21.2\\
          & iteration &10+14 & 10+15 & 6+14 & 6+12 & \textbf{5+9}\\ \midrule[0.3mm]
          \multirow{4}{*}{\rotatebox{90}{Picca.}} &$\Delta \bar{\text {deg}}$ & 5.12 & 5.19 & 3.93 & 4.08 & \textbf{3.82}\\
          & $\Delta \hat{\text {deg}}$ &2.02 & 2.34 & 1.81 & 1.83 & \textbf{1.77}\\
          & runtime & \textbf{153.5} & 170.2 & 191.9 & 180.6 & 173.4\\
          & iteration &10+16 & 10+19 & 6+21 & 6+19 & \textbf{6+16}\\ \midrule[0.3mm]
          \multirow{4}{*}{\rotatebox{90}{NYC Lib}} &$\Delta \bar{\text {deg}}$ & 2.71 & 2.66 & 2.63 & 2.63 & \textbf{2.56}\\
          & $\Delta \hat{\text {deg}}$ &1.37 & 1.30 & 1.24 & 1.20 & \textbf{1.17}\\
          & runtime & \textbf{2.5} & 2.6 & 4.5 &4.2 &3.7\\
          & iteration &10+14 & 10+15 & 6+14 & 6+11 & \textbf{6+9}\\ \midrule[0.3mm]
          \multirow{4}{*}{\rotatebox{90}{P.D.P}} &$\Delta \bar{\text {deg}}$ & 4.1 & 3.99 & 3.73 & 3.77 & \textbf{3.64}\\
          & $\Delta \hat{\text {deg}}$ &2.07 & 2.09 & 1.93 & 1.85 & \textbf{1.80}\\
          & runtime & \textbf{2.8} & 3.1 & 3.5 & 3.1 & 2.9\\
          & iteration &10+9 & 10+13 & \textbf{6+3} & 6+7 & 5+6\\ \midrule[0.3mm]
          \multirow{4}{*}{\rotatebox{90}{Rom. For.}} &$\Delta \bar{\text {deg}}$ & 2.66 & 2.69 & 2.62 & 2.65 & \textbf{2.60}\\
          & $\Delta \hat{\text {deg}}$ &1.58 & 1.57 & \textbf{1.37} & 1.44& {1.39}\\
          & runtime & \textbf{8.6} & 11.4 & 8.8 & 9.2 & 8.6\\
          & iteration &10+9 & 10+17 & 6+8 & 5+9 & \textbf{5+7}\\ \midrule[0.3mm]
          \multirow{4}{*}{\rotatebox{90}{T.o.L}} &$\Delta \bar{\text {deg}}$ & 3.42 & 3.41 & 3.16 & 3.23 & \textbf{3.03}\\
          & $\Delta \hat{\text {deg}}$ &2.52 & 2.50 & 2.20 & 2.12 & \textbf{1.98}\\
          & runtime & 2.6 & \textbf{2.4} & 2.7 & 2.5 & \textbf{2.4}\\
          & iteration &10+8 & 10+12& 6+7 & 6+7 & \textbf{5+6}\\ \midrule[0.3mm]
          \multirow{4}{*}{\rotatebox{90}{Uni. Sq.}} &$\Delta \bar{\text {deg}}$ & 6.77 & 6.77 & 6.54 & 6.57 & \textbf{6.24}\\
          & $\Delta \hat{\text {deg}}$ &3.66 & 3.85 & \textbf{3.48} & 3.70 & 3.64\\
          & runtime & \textbf{5.0} & 5.6 & 5.7 & 5.5 & 5.4\\
          & iteration &10+32 & 10+47 & 6+21 & \textbf{6+15} & {5+20}\\ \midrule[0.3mm]
          \multirow{4}{*}{\rotatebox{90}{Yorkm.}} &$\Delta \bar{\text {deg}}$ & 2.6 & 2.45 & 2.47 & 2.45 & \textbf{2.37}\\
          & $\Delta \hat{\text {deg}}$ &1.59 & 1.53 & 1.45 & 1.47 & \textbf{1.33}\\
          & runtime & \textbf{2.4} & 3.3 & 3.9 & 3.7 & 3.1\\
          & iteration &10+7 & 10+9 & 6+7 & 5+9 & \textbf{5+7}\\ \midrule[0.3mm]

	\end{tabular}}
	\end{center}
\end{table}
We provide the accuracy in mean and median degree, runtime and iteration numbers in Table.~\ref{table:result}, where Ours-$l_2$ denotes the corresponding result using $l_2$ cost function instead of the original $\rho$ in our proposed algorithm. It is shown that our proposed algorithm achieves the superior performance on almost all of the dataset.

{\bf Accuracy}
We measure the accuracy of the methods using the mean and median degree error by convention. We notice that the result given by that using $l_2$ cost function bares slightly greater error than that with cost function $\rho$ in our original scheme. The reason can be due to the differences on both the weight computation and the influence function. As we mentioned before, the influence function of $l_2$ cost is unbounded such that, during the initialization the penalty on the measurement deviation is too harsh. In that case, some edges might get over-penalized in the sense of the `good' information gets underweighted in the optimization. Moreover, as $\rho$ in our scheme will lower the penalty term as the iteration goes on, the optimization will be better refined when the solve is sufficiently close to the solution set.  

{\bf Speed}
It can be seen that our proposed algorithm is slightly slower than the fastest approach IRLS on most of the datasets. As the plain-version IRLS processes the iteration without robostifier, it involves a lot fewer variables compared to the other approaches. Also it is worth to note that our algorithm takes the fewest iterations on almost all of the datasets, which greatly results from the weight initialization scheme such that IRLS is well intialized at the beginning, which leads to the faster convergence.
\begin{figure*}[!ht]
\begin{center}

\subfloat	[Sequence 00]	{
\includegraphics[width=.33\linewidth]{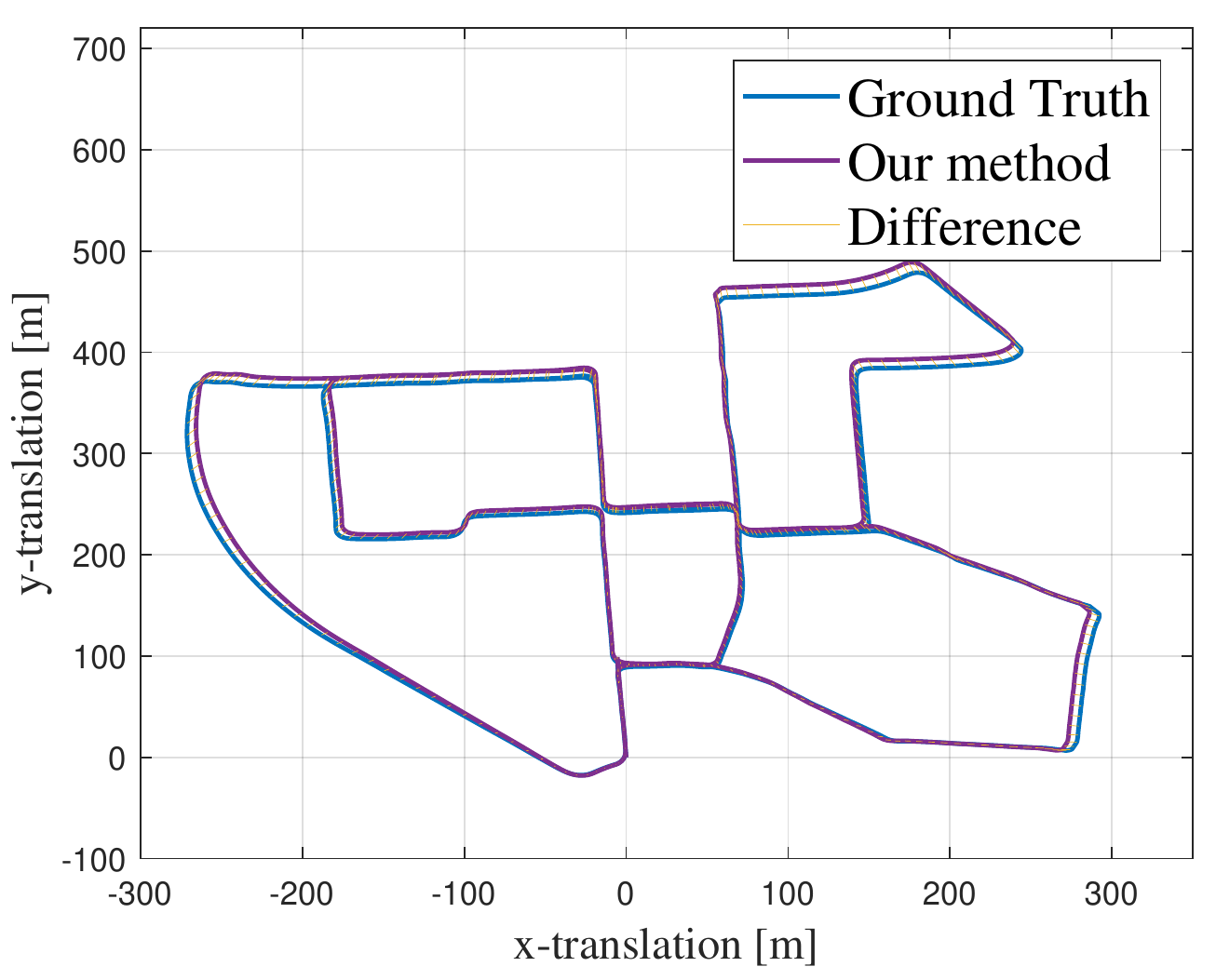}
}
\subfloat	[Sequence 02]	{
\includegraphics[width=.33\linewidth]{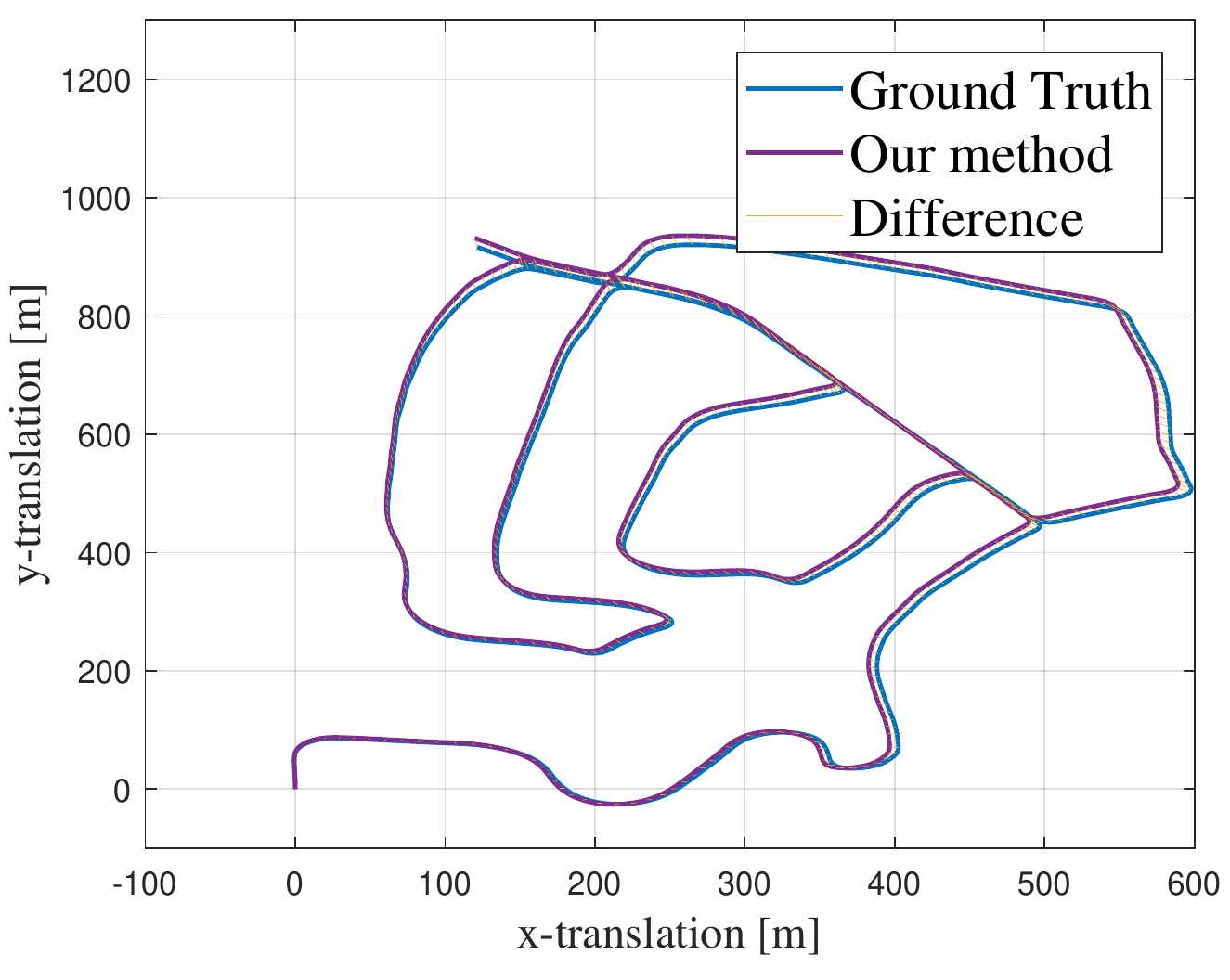}
}

\subfloat  [Sequence 05]    {
\includegraphics[width=0.33\linewidth]{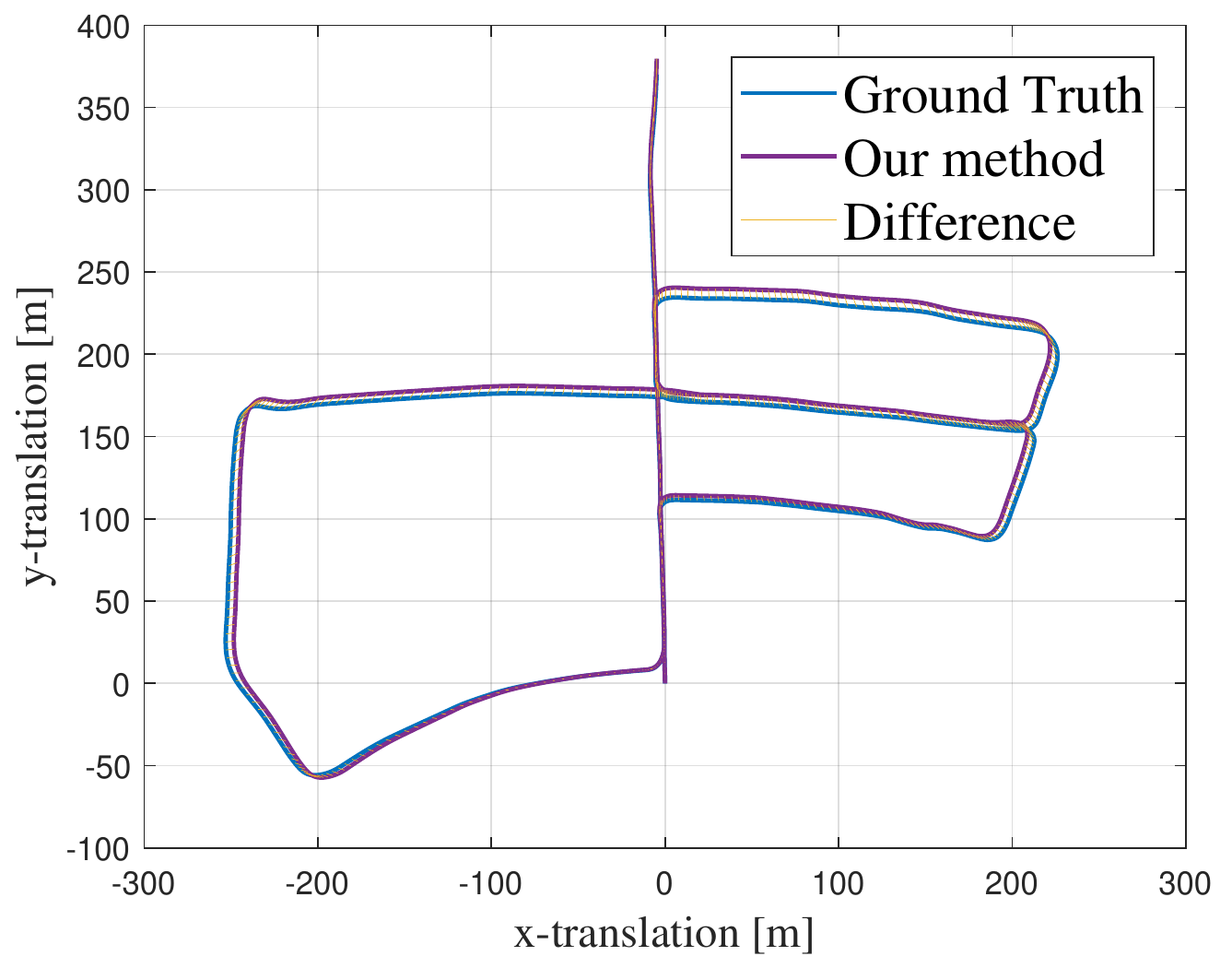}
}
\subfloat   [Sequence 09]   {
\includegraphics[width=0.32\linewidth]{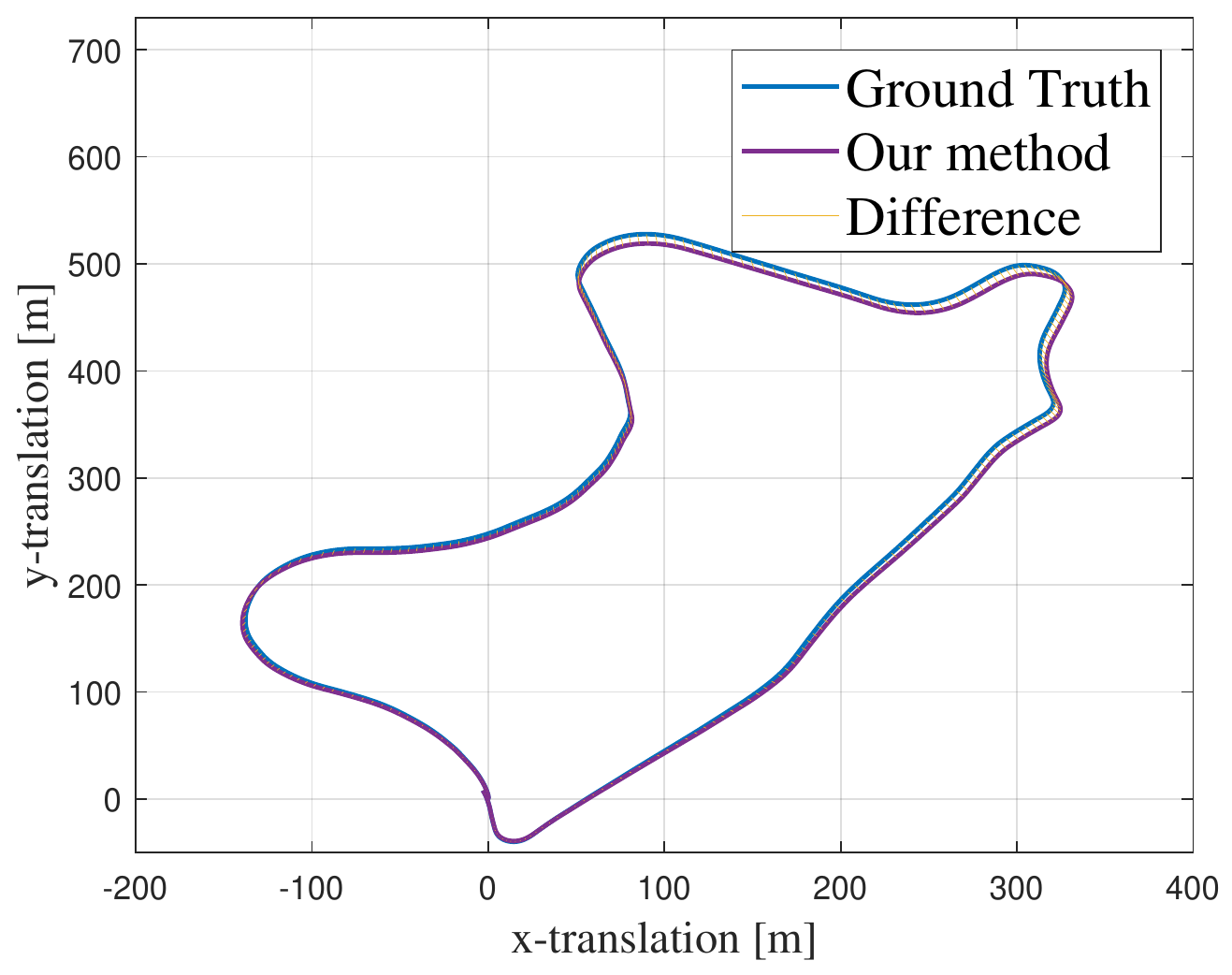}
}
\caption[Comparison with ORB-SfM on Seq.02, 08, 10 from KITTI Odometry.]{Our proposed approach shows a trajectory estimation of comparably high quality in KITTI dataset which contains large scale outdoor scenes. Specifically, our approach displays superior loop-closing capability on Seq.02, where many loops are present. It can be shown that the accumulated drifts are negligible through the trajectory estimation. Results on the rest of the dataset are provided in the supplementary materials.}
\label{fig:3seq}
\end{center}
\end{figure*}
\subsection{Qualitative Results}
To test the qualitative performance of our algorithm, we run experiments on VIO benchmark KITTI to further demonstrate the robustness of our proposed solver. Different from internet photo dataset like~\cite{wilson_eccv2014_1dsfm}, the pose-graph for KITTI contains a lot more small cycles but not many large-scale cycles. The local connectiviey is higher while the global connectivity is very small. To estimate the camera trajectory, we use standard SfM pipeline where we fix the camera orientation with our outputs to estimate the translation. 

\subsection{Ablation Study}
Figures of the ablation comparisons are provided in the supplementary materials, we state the main observations here. In the comparisons we use IRLS~\cite{chatterjee2018robust} as the baseline. We produce the synthetic data using perturbations from $0^{\circ}$ to $15^{\circ}$ uniformly on ground truth value of ~\cite{wilson_eccv2014_1dsfm}.
\subsubsection{Effects on Cost Choice}
\label{sec:ablationnorm}
We further conduct ablation study on the effects on the cost function choice with baseline as Robust IRLS~\cite{chatterjee2018robust} on synthetic data with perturbations on the measurements. It has been shown that our proposed algorithm with $l_1$, $l_2$ and Huber cost function all show superior accuracy and faster convergence over~\cite{chatterjee2018robust}.
\subsubsection{Effects on Cycle Consistency Enforcement}
We further test the cycle consistency constraints effect by conducting experiments with our solver with RANSAC, RANSAC+cycle constraints, and solely with the cycle constraints, with baseline~\cite{chatterjee2013efficient}. It has been shown that, it does not show much difference beween with RANSAC+cycle constraints vs solely with cycle constraints. 
\label{sec:ablationconsistency}

\section{Conclusion}
In this paper, we propose a robust framework to solve multiple rotation averaging problem, especially in the cases that a significant amount of noisy measurements are present. By introducing the $\epsilon$-cycle consistency term into the solver, we enable the robust initialization scheme to be implemented into the IRLS solver. Instead of conducting the costly edge removal, we implicitly constrain the negative effect of erroneous measurements by weight reducing, such that IRLS failures caused by poor initialization can be effectively avoided. Combined with our novel cost function, our proposed method outperforms state-of-the-art approaches significantly in both accuracy and efficiency.

\title{\LARGE{\bf{Supplementary Materials}}}
\maketitle

In this supplementary work, we will unfold our discussions in the main paper in more details. In the following, we include properties and results of rotation groups and quaternions, along with their representations on the unit 3-sphere (Fig.~\ref{fig:s2sphere} in the main paper) in \S\ref{sec:quat}. We conclude the work by providing the additional results of the experiments, where in \S\ref{sec:expquant} and \S\ref{sec:expqual} we give the quantitative and qualitative results on Photo Tourism~\cite{wilson_eccv2014_1dsfm} and KITTI~\cite{Geiger2012CVPR} respectively, detailed analysis of the ablation study is provided in \S\ref{sec:ablationresult}.
\section{Rotations, Quaternions and 3-sphere}
\label{sec:quat}
To generalize our analysis and make the discussions concise, we will adopt the notations $\{\mathbf{R}_1, \mathbf{R}_2, \cdots \}$ where $\mathbf{R}_i \in \mathbb{SO}(3)$ instead of $\{\lambda (1), \lambda (2), \cdots \}$ in the main paper. Similarly, $\{\mathbf{R}_{ij}\}$ denotes the same transformations as $\sigma (i,j)$. As all rotations can be uniquely (up to a scale) represented by unit quaternions, we further exploit the quaternion representation of $\{\mathbf{R}_i\}$. Specifically, let a unit quaternion $\mathbf{q}_i$ denote the same rotation as $\mathbf{R}_i$. Recall that
\begin{equation}
    \mathbf{q}_i = \mathbf{s}_i + \alpha _i \mathbf{i} + \beta _i \mathbf{j} + \gamma _i \mathbf{k}, 
\end{equation}
where
\begin{align}
    \mathbf{i}^2 = &\mathbf{j}^2 = \mathbf{k}^2 = \mathbf{ijk} = -1,\\
    &\mathbf{ij} = \mathbf{-ji} = \mathbf{k},\\
    &\mathbf{jk} = \mathbf{-kj} = \mathbf{i},\\
    &\mathbf{ki} = \mathbf{-ik} = \mathbf{j},\\
    \mathbf{s}_i^2 + &\alpha _i^2 + \beta _i^2 + \gamma _i^2 = 1.
\end{align}
We will use the concrete notation $\mathbf{q}=[\mathbf{s}, \mathbf{v}]$ henceforth, where $\mathbf{s}$ denotes the real part of $\mathbf{q}$ and $\mathbf{v} = [\alpha, \beta, \gamma]$ denotes the imaginary part. Let $\mathbf{q}_1$ denotes an arbitrary rotation $\mathbf{R}_1 \in \mathbb{SO}(3)$, we have that there exists a set of basis $\{e_1^i\}$ such that $\mathbf{q}_1=\sum a_i e_1^i$ for $a_i \in \mathbb{R}, \forall i$. Then for $\mathbf{q}_{12}$ which represents the relative rotation $\mathbf{R}_{12}$, we thus have $\mathbf{R}_2 = \mathbf{R}_{12}\mathbf{R}_1$ which in quaternionic form is 
\begin{equation}
\label{eq:q2}
    \mathbf{q}_2 = \mathbf{q}_{12} \mathbf{q}_1 \mathbf{q}_{12}^{-1} = \mathbf{q}_{12} \big(\sum a_i e_1^i \big) \mathbf{q}_{12}^{-1}.
\end{equation}
While Eq.~\ref{eq:q2} cannot be directly expanded, it suffices to show that $\mathbf{q}_2$ can be spanned with the basis rotated by $\mathbf{q}_{12}$. Let $\{e_{12}^i\}$ denotes the basis of $\mathbf{q}_{12}$, it immediately follows that the rotated basis set from $\{e_{1}^i\}$ is that, $e_2^i = e_{12}^i e_1 (e_{12}^i)^{-1} $. Substituting into Eq.~\ref{eq:q2}
\begin{equation}
    \mathbf{q}_2 =\sum a_i e_{12}^i e_1 (e_{12}^i)^{-1} =\sum a_i e_2^i .
\end{equation}
The generalized form of the equation above is thus
\begin{equation}
\label{eq:generalq}
    \mathbf{q}_n = \sum a_i e_{n-1,n} e_{n-2, n-1} \cdots e_{12}^i e_1 (e_{12}^i)^{-1} \cdots e_{n-2, n-1}^{-1} e_{n-1,n} ^{-1} .
\end{equation}
Eq.~\ref{eq:generalq} provides the equivalence between the progressive rotation multiplication and the summation with respect to the transformed quaternion basis. Now we can change Def.~\ref{def:epsconsistency} into that with quaternion form, which essentially denotes the geodesic curve length on the surface of unit 3-sphere $\mathcal{S}^3$. Recall that we define the $\epsilon$-cycle consistency as
\begin{equation}
\label{eq:epsrotation}
     d(\sigma(i,i_1)\cdot \prod_{j=1}^{m_k -1} \sigma({i_j}, i_{j+1}) \cdot \sigma(i_{m_k},i),\mathbb{I}_\sigma)\leq \epsilon .
\end{equation} 

\begin{lemma}
Let $\mathbf{q}_{ij}$ denote the relative rotation as defined in Eq.~\ref{eq:epsrotation}, \ie, $\mathbf{q}_{ij} = \sigma (i,j), \forall i,j$. Then given $\mathbf{q}_0$, there exists an angle of rotation $\theta$ such that
\begin{equation}
    d_{\angle}(\mathbf{q}_n, \mathbf{q}_0) \leq \theta,
\end{equation}
is equivalent with Eq.~\ref{eq:epsrotation}, where $\mathbf{q}_n$ is the quaternion after $n$ rotations from $\mathbf{q}_0$.
\end{lemma}
The proof follows by realizing that any quaternion can be written into the rotation axis and an rotation angle $\theta$. It immediately follows that the line determined by the rotation axis is invariant under the rotation angle. Assume without loss of generality that $\mathbf{q}_n$ has the same rotation axis with $\mathbf{q}_0$, it then suffices to show that the pure quaternion of  $\mathbf{q}_0$ after transformation by progressive $\mathbf{q}_{ij}$ is also a pure quaternion. Consider $\mathbf{q}_j = \mathbf{q}_{ij} \mathbf{q}_i \mathbf{q}_{ij}^{-1}$, we thus have
\begin{equation}
\begin{aligned}
    \mathbf{q}_{ij} \mathbf{q}_i \mathbf{q}_{ij}^{-1} &= \mathbf{q}_{ij} (\mathbf{s}_i+\mathbf{v}_i) \mathbf{q}_{ij}^{-1}\\
    &= \mathbf{q}_{ij} \mathbf{s}_i\mathbf{q}_{ij}^{-1}+\mathbf{q}_{ij}\mathbf{v}_i \mathbf{q}_{ij}^{-1}\\
    &=\mathbf{s}_i + \mathbf{q}_{ij}\mathbf{v}_i \mathbf{q}_{ij}^{-1},
    \end{aligned}
\end{equation}
and the lemma follows.

\section{Additional Experimental Results}
\label{sec:exp}
In this section we first provide additional quantitative results on the Photo Tourism Dataset~\cite{wilson_eccv2014_1dsfm} in \S\ref{sec:expquant}. We have tested our proposed approach with variations, \eg, with conventional cost functions and RANSAC preprocessings. We demonstrate the efficiency of our proposed approach by outperforming the state-of-the-arts approaches on both speed and accuracy. We then present the qualitative results on the KITTI dataset~\cite{Geiger2012CVPR}, where the accurate camera trajectories have demonstrated the robustness of our proposed approach in consecutive image sequence, \ie, smaller cycles are ubiquitous in the pose-graph while large cycles are normally absent. Furthermore, we conduct ablation studies on the effects of cycle constraints and cost function choices in~\S\ref{sec:ablationresult} to conclude our discussions.
\subsection{Full Quantitative Results on~\cite{wilson_eccv2014_1dsfm}}
\label{sec:expquant}
In Table~\ref{table:fulltable}, we provide the performance on the Photo Tourism Dataset~\cite{wilson_eccv2014_1dsfm}, compared with the original IRLS~\cite{chatterjee2013efficient}, the Robust IRLS~\cite{chatterjee2018robust}, MPLS~\cite{lerman2019robust}. Our proposed approach outperforms the state-of-the-art methods by both speed and accuracy on most of the datasets.

We have conducted the experiments with different cost functions and with different initialization schemes. In Table~\ref{table:fulltable}, `Ours-$l_2$', `Ours-$l_1$' and `Ours-$l_{\frac{1}{2}}$' represent our proposed approach with $l_2$, $l_1$ and $l_{\frac{1}{2}}$ cost functions, respectively. It can be shown that though that with $l_{\frac{1}{2}}$ cost function achieves a better accuracy than the other two cost functions, it takes notably longer processing time and more iterations as well. Moreover, on large-scale dataset, the tradeoff is quite inefficient. For example, on the Piccadilly dataset, which contains more than 2000 images, the $l_{\frac{1}{2}}$ takes more than 15\% runtime compared with the $l_2$ optimization scheme while only providing negligible edge on accuracy advantage. Meanwhile out proposed method with exponential cost function runs slightly longer than~\cite{chatterjee2013efficient}, but improves the accuracy tremendously.

In addition, we have tested our approach with `RANSAC' only preprocessing and `RANSAC+cycle constraints' and it can be shown that with our proposed cycle constraint on the erroneous relative rotations, the accuracy has been improved substantially. It should be noted that all of the state-of-the-art approaches we are comparing our approach with have implemented RANSAC iterations before the rotation averaging to filter the measurements. It is shown that although `Ours-RANSAC' outperforms the other approaches on a small scale on several datasets, the accuracy is rather low compared with `Ours-R+cycle' and `Ours'. It also should be noted that, without our proposed initialization schemes, the optimization requires more iterations to be successfully initialized. Furthermore, the comparison between `Ours-R+cycle' and `Ours' validates the effects of
our proposed enforcement of the cycle constraints. It can be shown that the two mostly perform similarly on most of the datasets. The fact that the additional RANSAC does not improve the accurac demonstrates the robustness of our proposed approach against the measurement outliers.

\subsection{Full Qualitative Results on~\cite{Geiger2012CVPR}}
\label{sec:expqual}
\begin{figure*}[!ht]
\begin{center}

\subfloat	[Sequence 00]	{
\includegraphics[width=.33\linewidth]{seq00.pdf}
}
\subfloat	[Sequence 01]	{
\includegraphics[width=.33\linewidth]{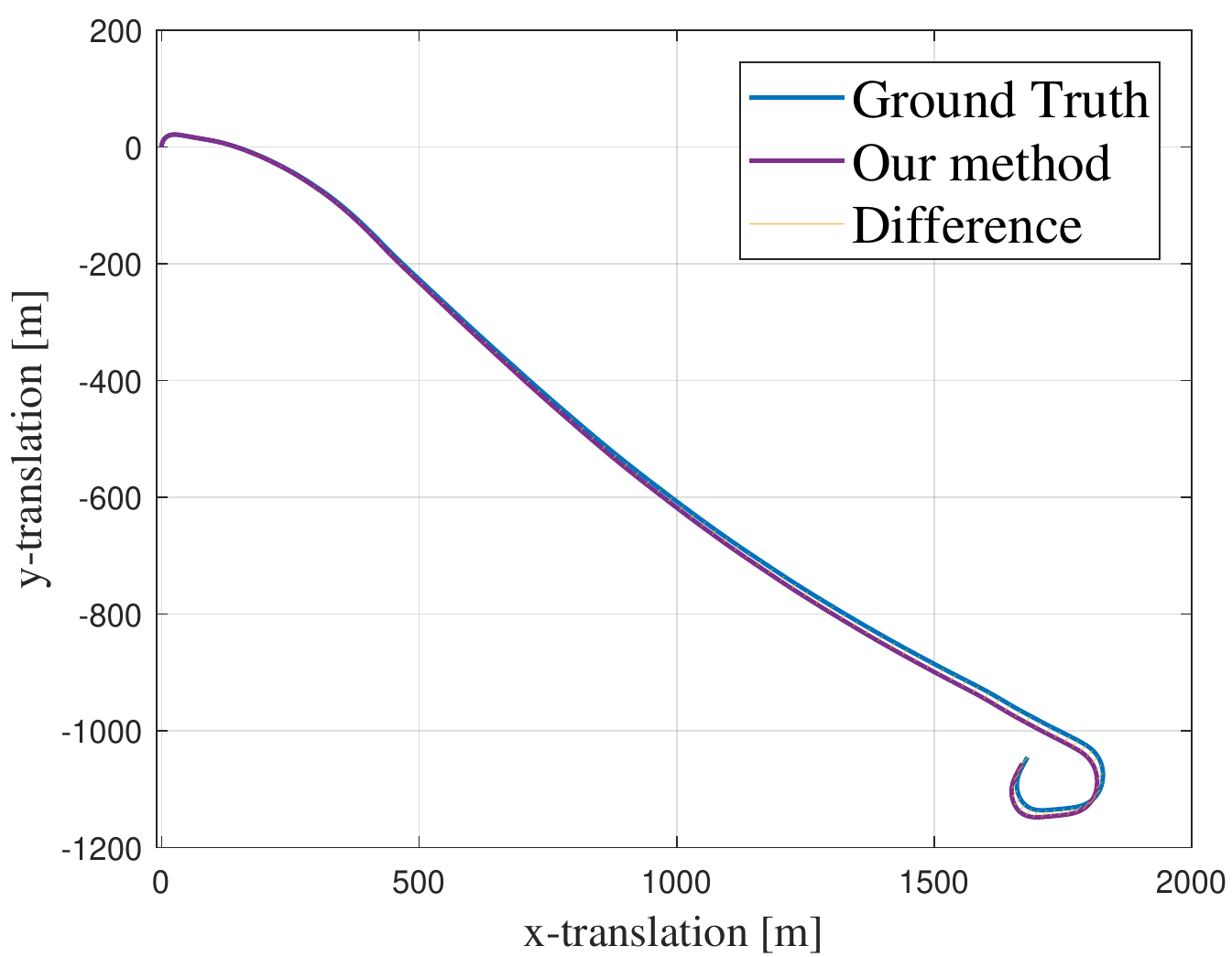}
}
\subfloat	[Sequence 02]	{
\includegraphics[width=.33\linewidth]{seq02.pdf}
}

\subfloat	[Sequence 03]	{
\includegraphics[width=.33\linewidth]{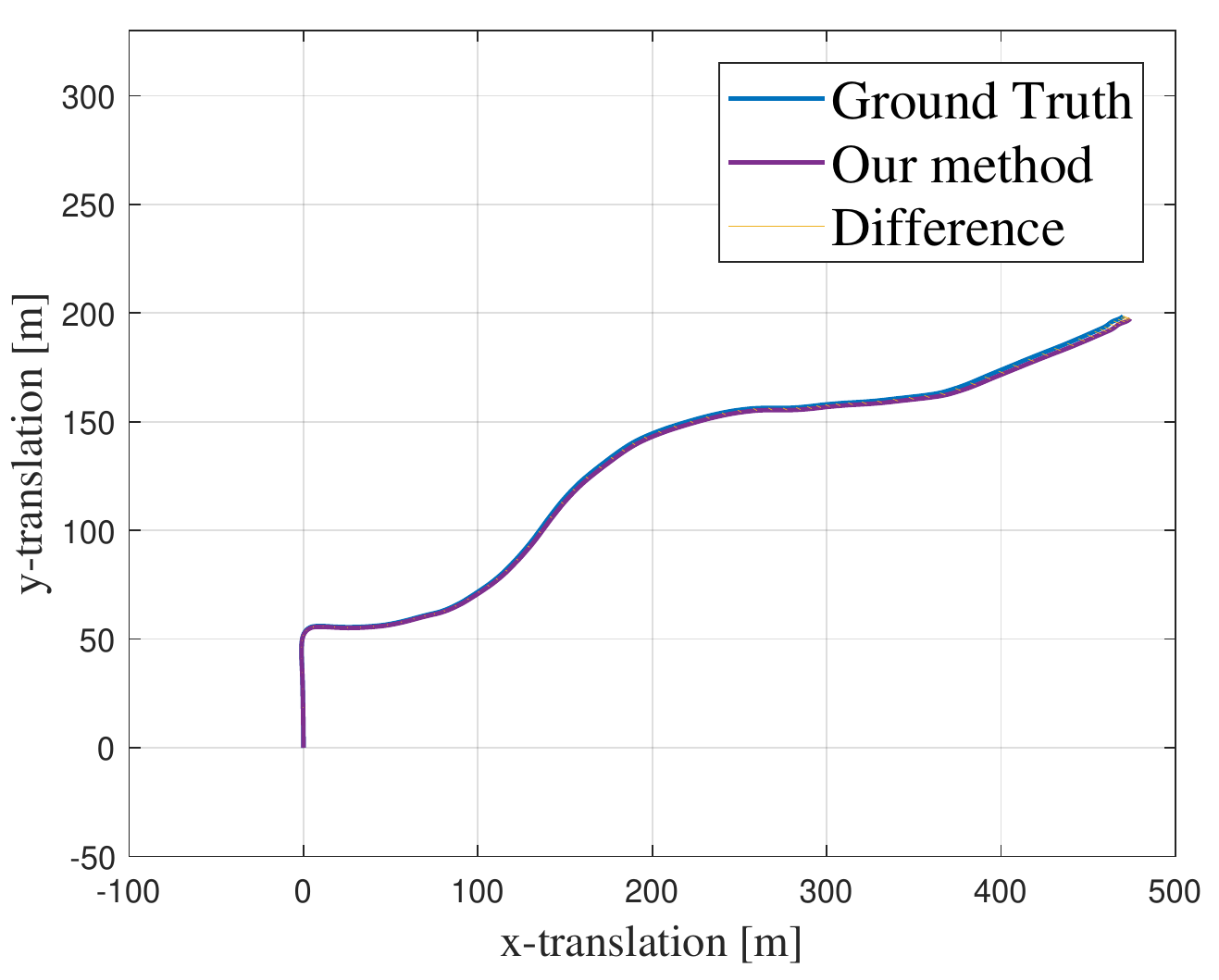}
}
\subfloat	[Sequence 04]	{
\includegraphics[width=.33\linewidth]{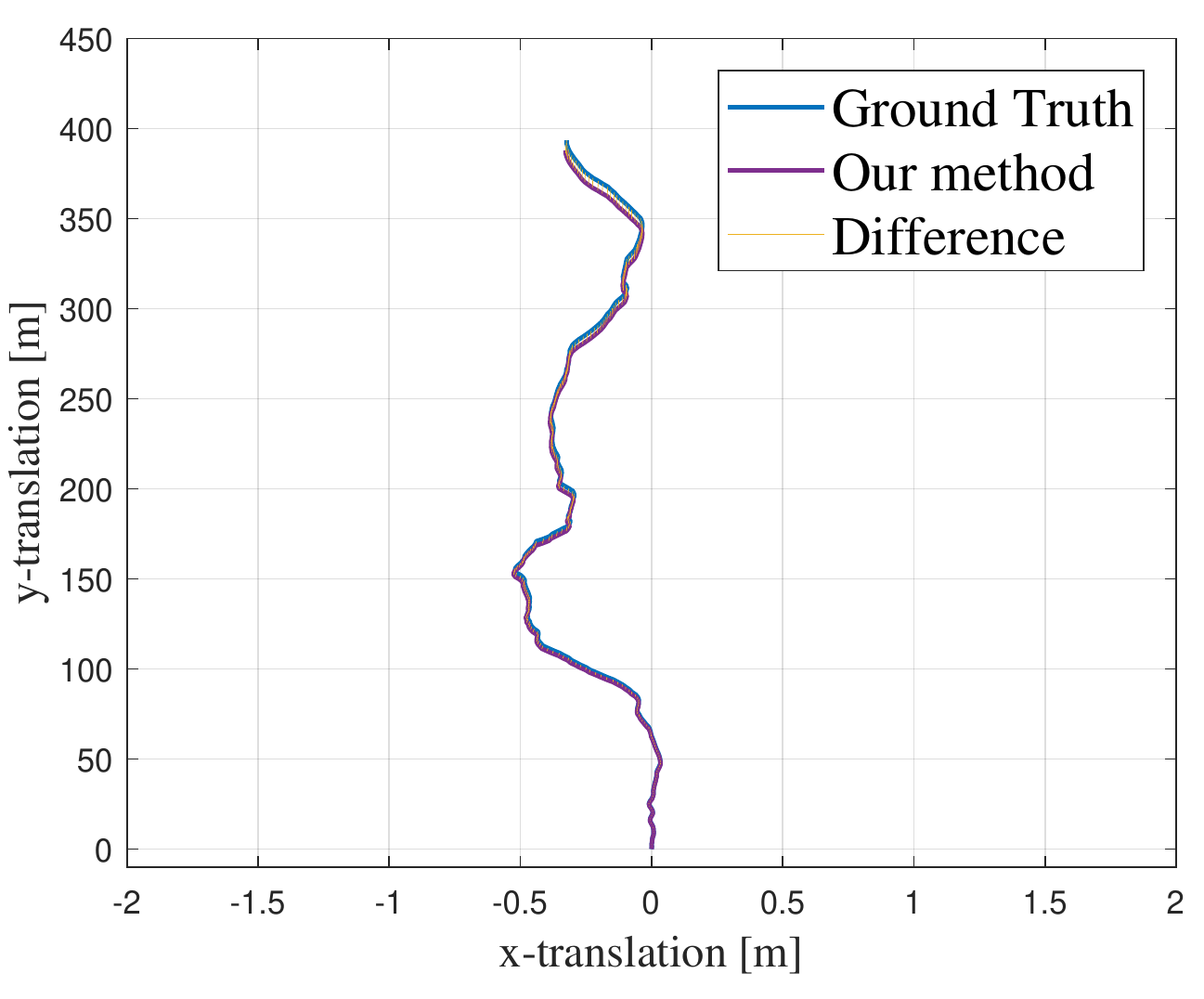}
}
\subfloat  [Sequence 05]    {
\includegraphics[width=0.33\linewidth]{seq05.pdf}
}

\subfloat	[Sequence 06]	{
\includegraphics[width=.33\linewidth]{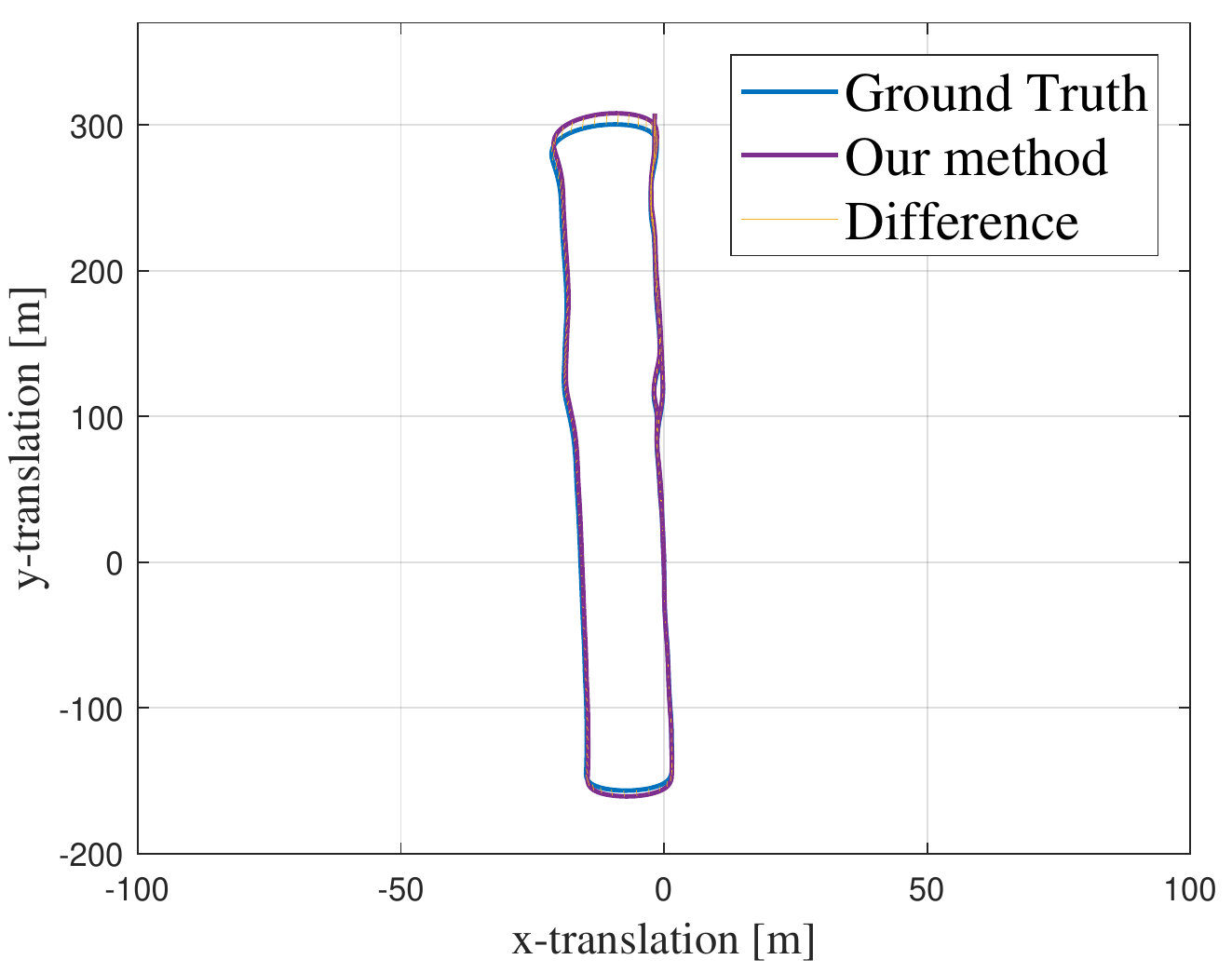}
}
\subfloat	[Sequence 07]	{
\includegraphics[width=.33\linewidth]{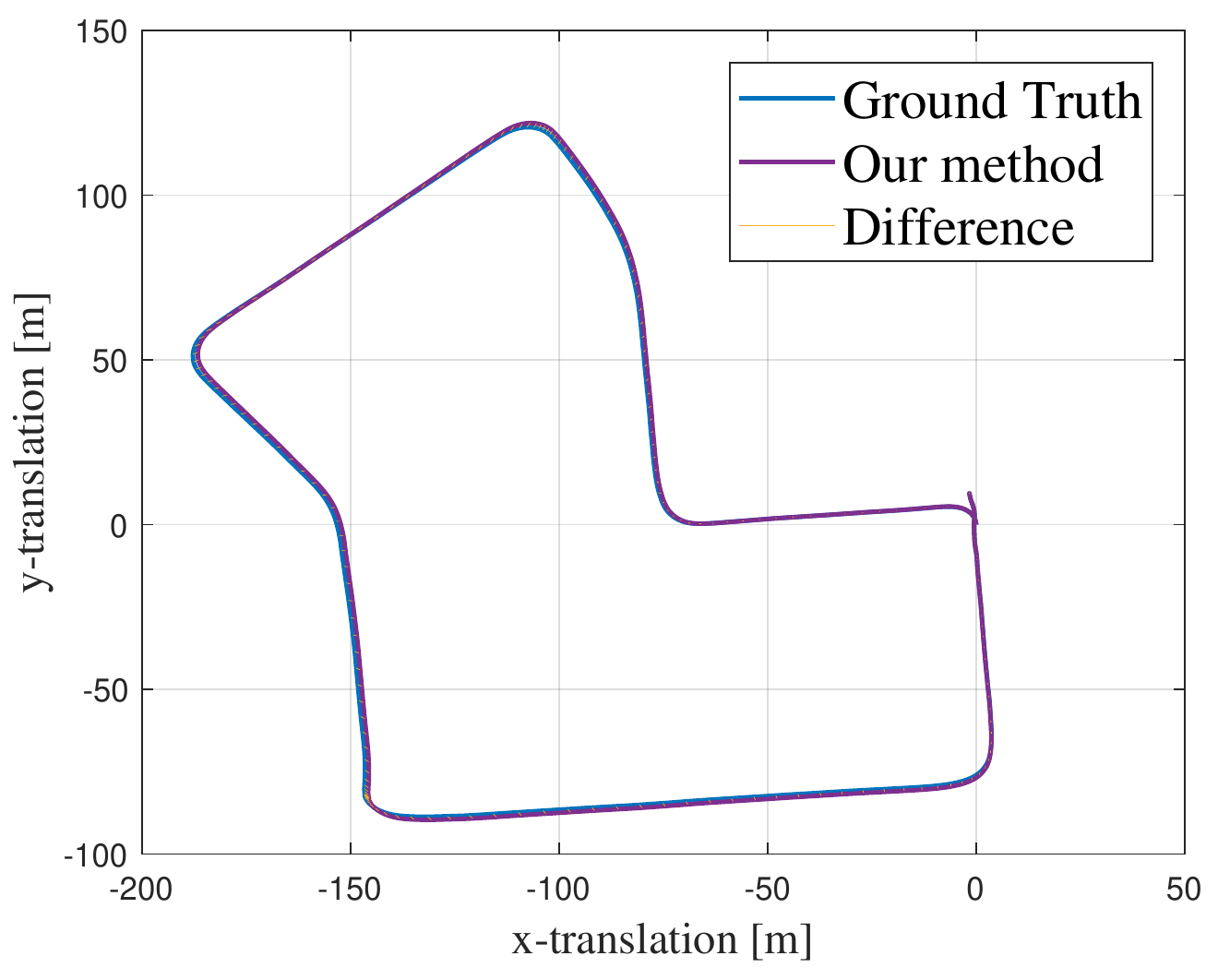}
}\subfloat	[Sequence 08]	{
\includegraphics[width=.33\linewidth]{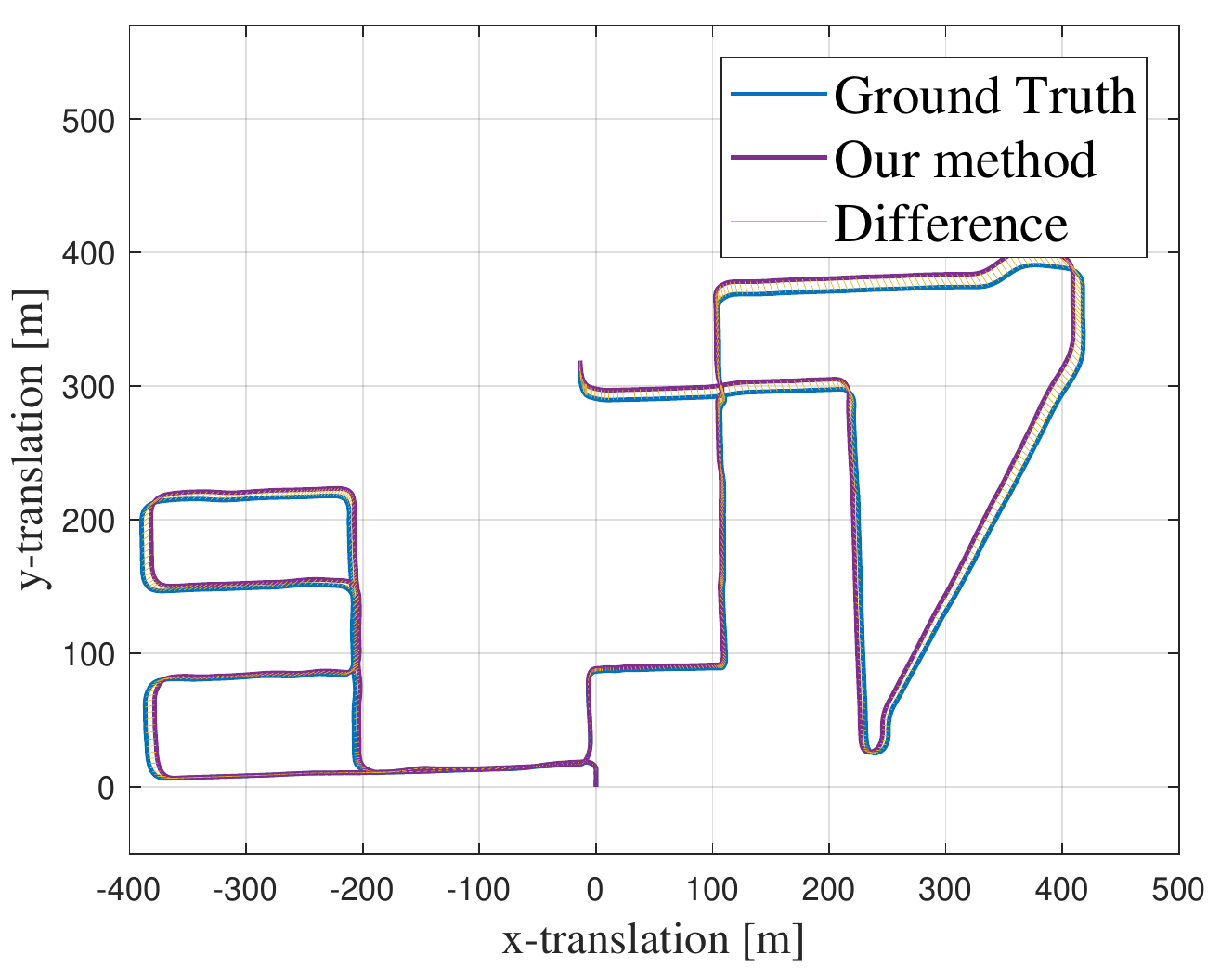}
}

\subfloat   [Sequence 09]   {
\includegraphics[width=0.32\linewidth]{seq09.pdf}
}
\subfloat	[Sequence 10]	{
\includegraphics[width=.33\linewidth]{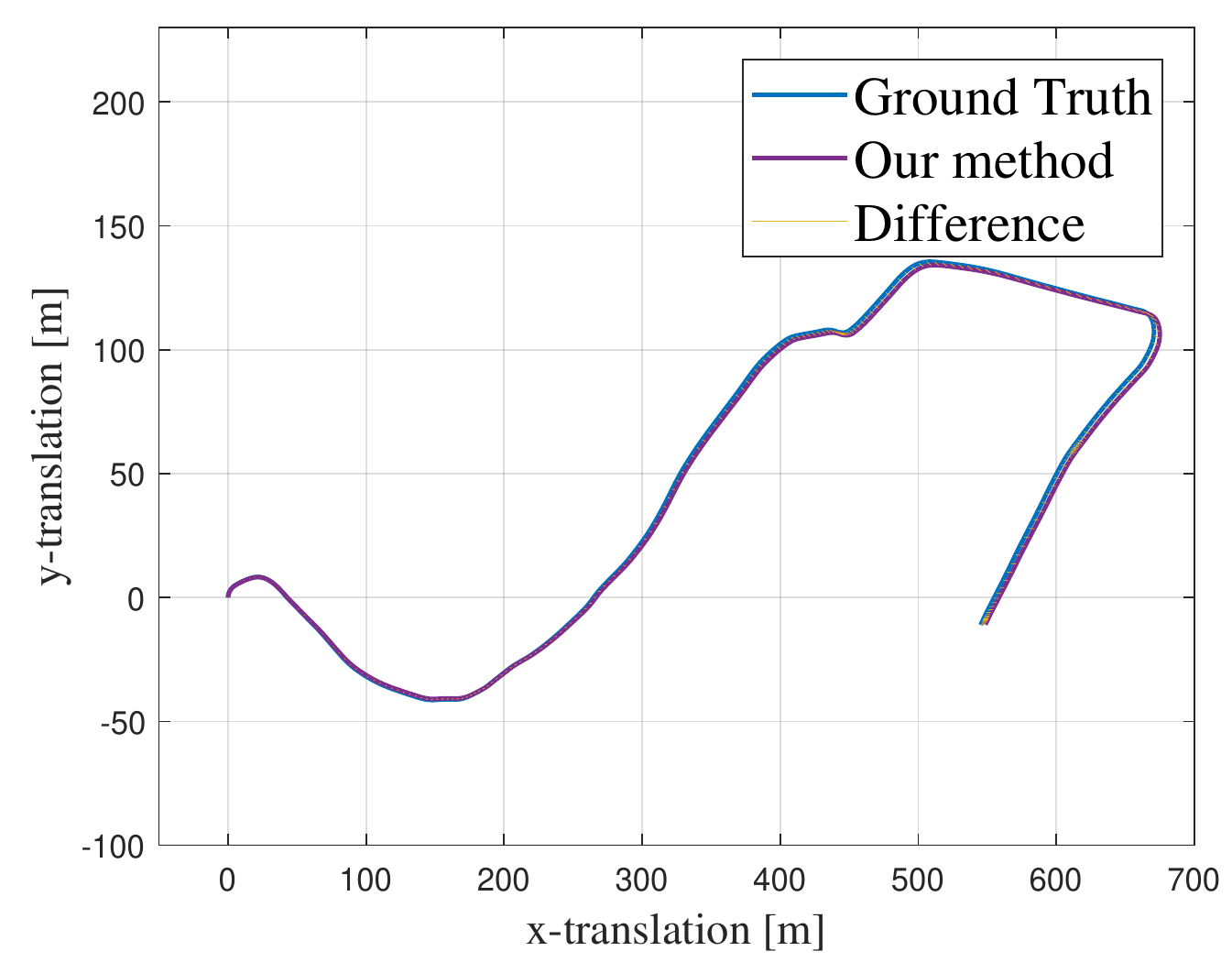}
}
\caption{Our proposed approach shows a trajectory estimation of comparably high quality in KITTI dataset which contains large scale outdoor scenes.}
\label{fig:allkitti}
\end{center}
\end{figure*}

In Fig.~\ref{fig:allkitti}, we provide the camera trajectories given by our proposed approach on the KITTI dataset~\cite{Geiger2012CVPR}. In details, we first solve for the camera rotations and then keep the rotations fixed and compute the camera translation with conventional bundle adjustment~\cite{ceres-solver}.

In the experiments it can be observed that, our proposed approach has shown consistent robustness when dealing with consecutive image sequences. As translation estimation includes the approximation of unknown scale parameter, deviation in rotation estimation can be amplified catastrophically. The accurate camera trajectories thus demonstrate the high accuracy delivered by our rotation averaging scheme. Specifically, most of the sequences do not contain large cycles in the viewgraph, in such scenarios the enforcement of cycle consistency tends to yield more loose constraints. However, since there exists abundant smaller cycles in the viewgraph, the constraints will be denser in the subspace, thus demonstrate the practicality and universality of our proposed approach. 

\subsection{Ablation Study}
\label{sec:ablationresult}
\begin{figure}[th!]
    \centering
   \subfloat	[Noise=10\%]	{
\includegraphics[width=.5\linewidth]{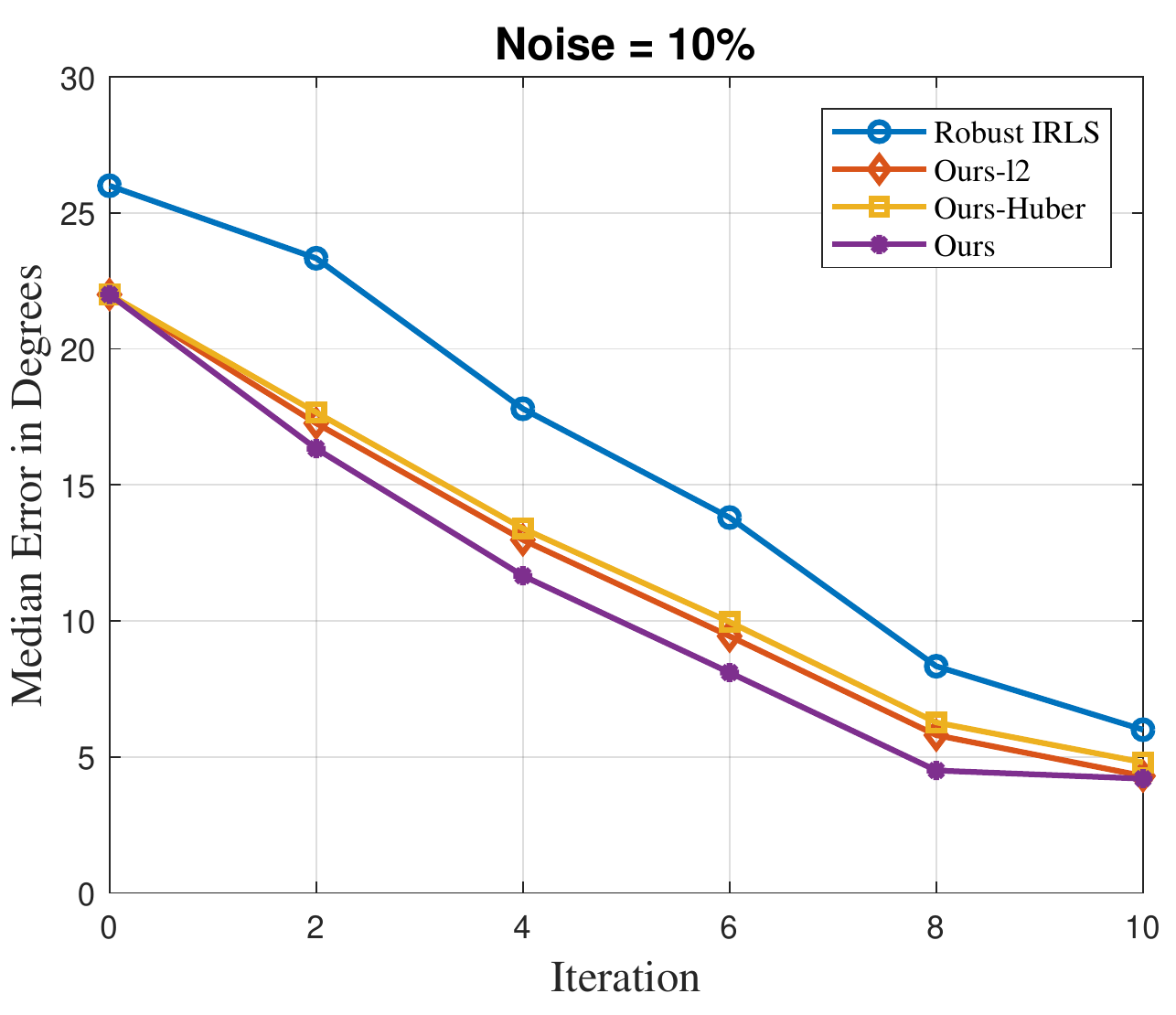}
}
\subfloat	[Noise = 20\%]	{
\includegraphics[width=.5\linewidth]{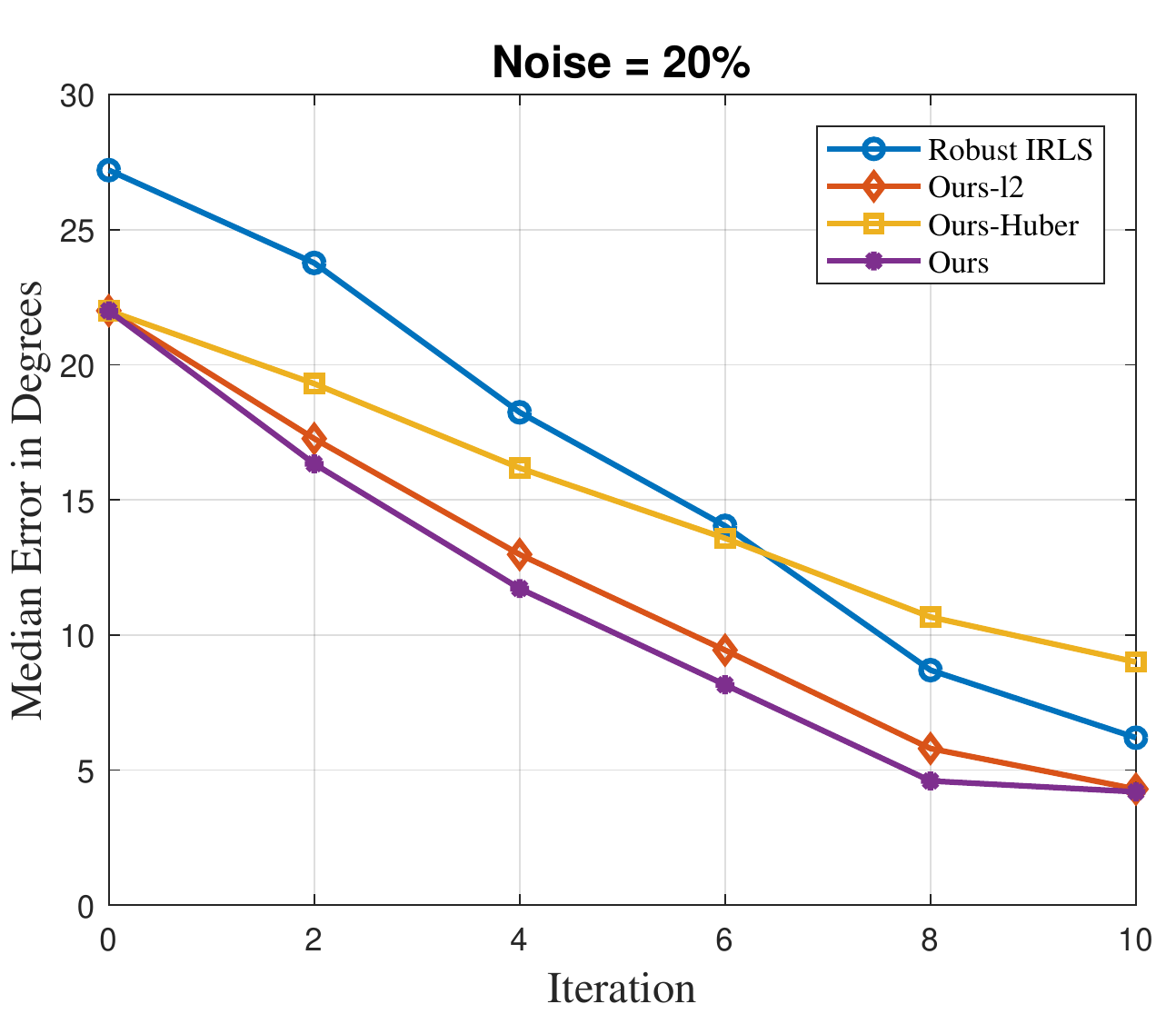}
}

\subfloat  [Noise = 30\%]    {
\includegraphics[width=0.5\linewidth]{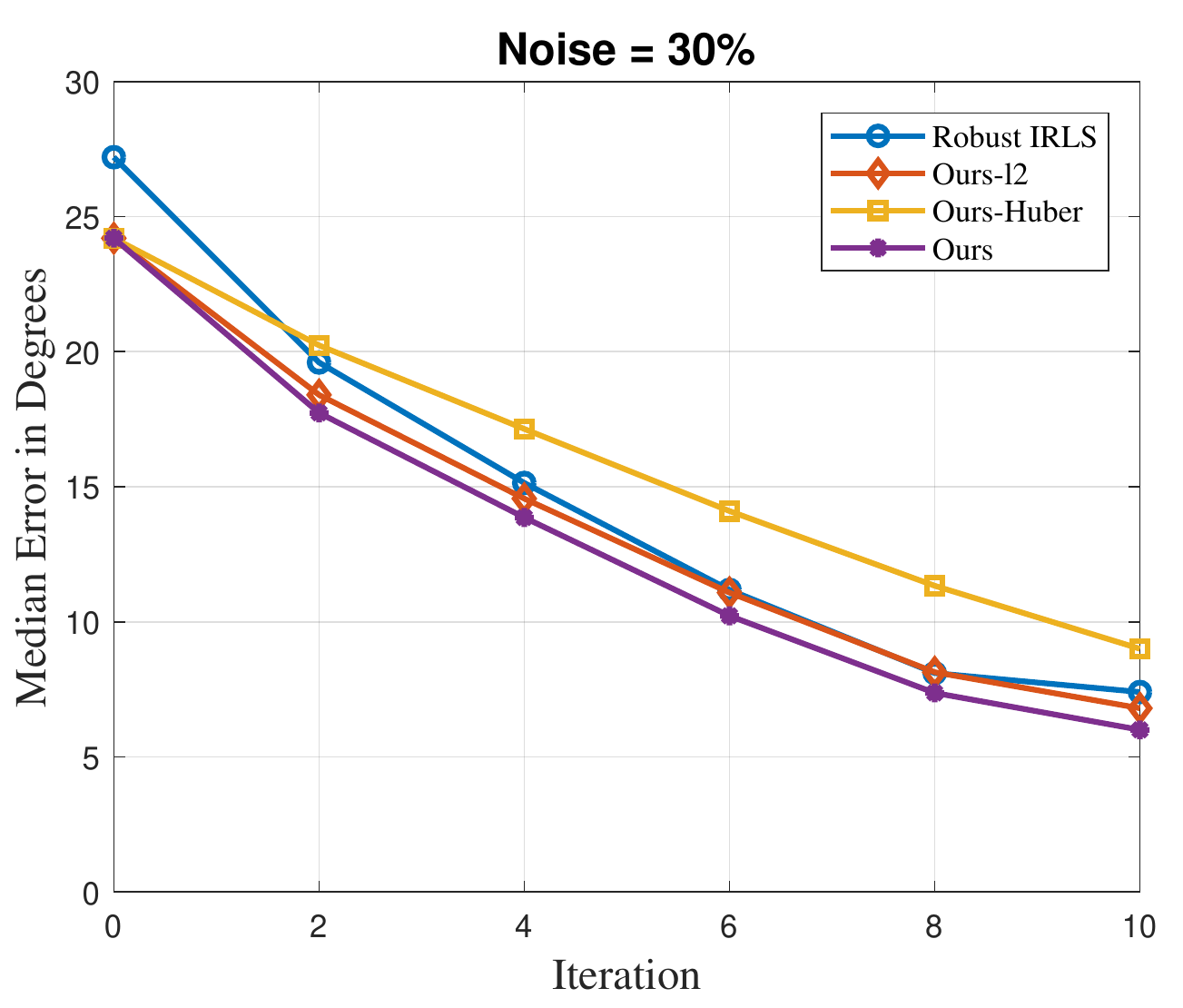}
}
\subfloat  [Noise = 40\%]    {
\includegraphics[width=0.5\linewidth]{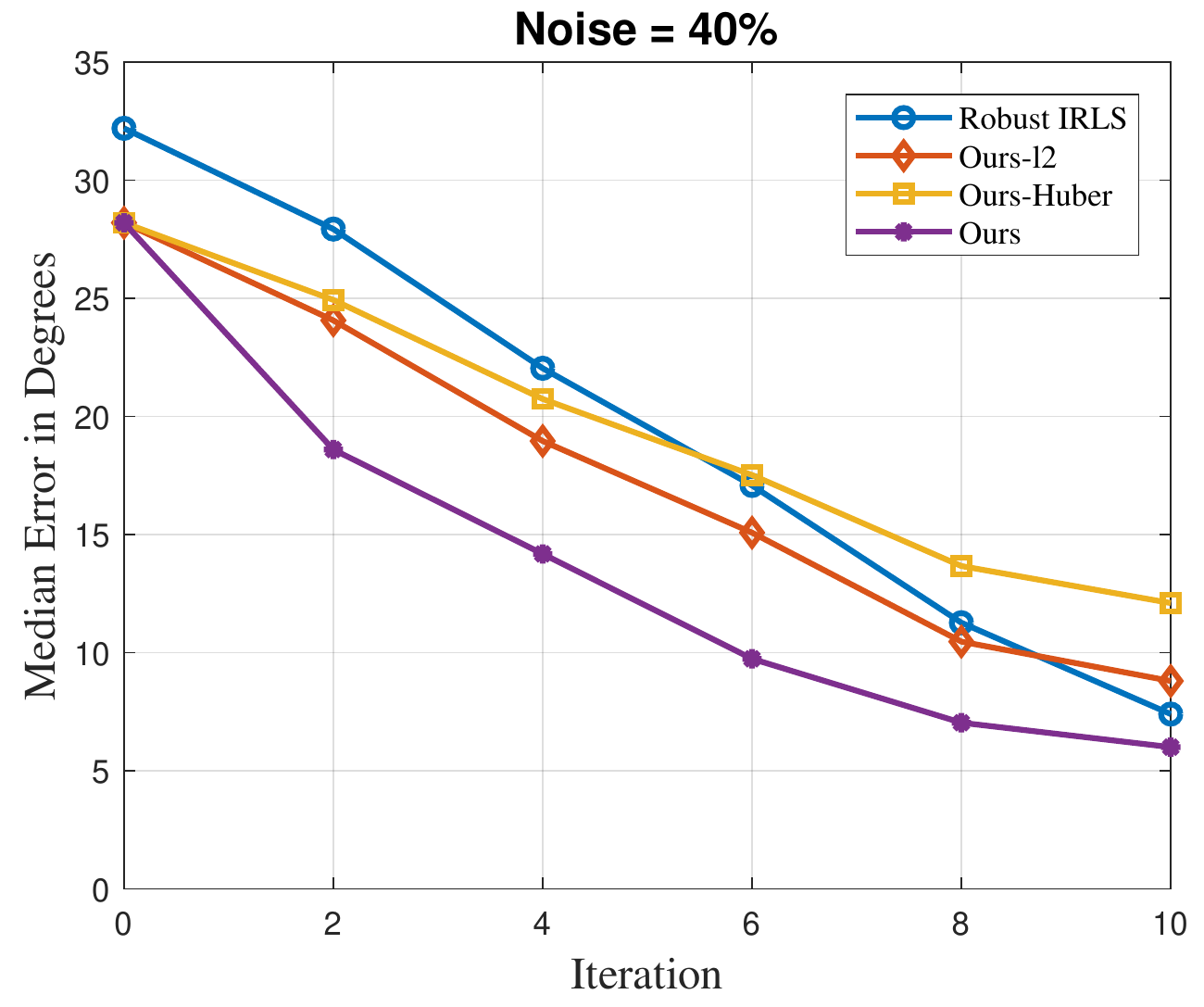}
}
    \caption{Convergence performances with different cost function under different noise levels.}
    \label{fig:convfig}
\end{figure}
In the following experiments, we randomly sample ground truth values of 100 camera rotations from the Piccadilly dataset from~\cite{wilson_eccv2014_1dsfm} and compute the relative rotations with them. We then randomly select 10\%, 20\%, 30\%, 40\% of the relative rotations to add in noise uniformly. We first analyze the convergence performance with different cost functions, followed by the robustness analysis with different measurement filtration schemes. For each experiment, we record the result by repeating the experiment with same setting for 10 times and report the mean values.
\subsubsection{Cost Function}
With the synthetic data we test the convergence performances of Robust IRLS~\cite{chatterjee2018robust}, Ours-$l_2$, Ours-Huber and Ours with different noise and outlier levels. In all the experiments, we perturb the corresponding proportion of the ground truth relative rotations by $5^{\circ}$.

The convergence performances are given in Fig.~\ref{fig:convfig}. It can be observed that as noise level and outliers increase, our propose solver shows strong robustness and displays stable and fast convergence. While our approach with different cost functions show similar convergence property when the noise level is low, as 40\% of the edges have add-in noise, the Huber cost function shows the slowest convergence. In our experiments, that with Huber cost always requires more than 15 iterations to converge when the outliers are relatively high. It also worth to point out that when a high noise level is present, the properties of our cost function can be shown more clearly, that the penalty is high at the beginning of the optimization and slows down near the optimal point.

\begin{figure}[th!]
    \centering
   \subfloat	[Noise=10\%]	{
\includegraphics[width=.5\linewidth]{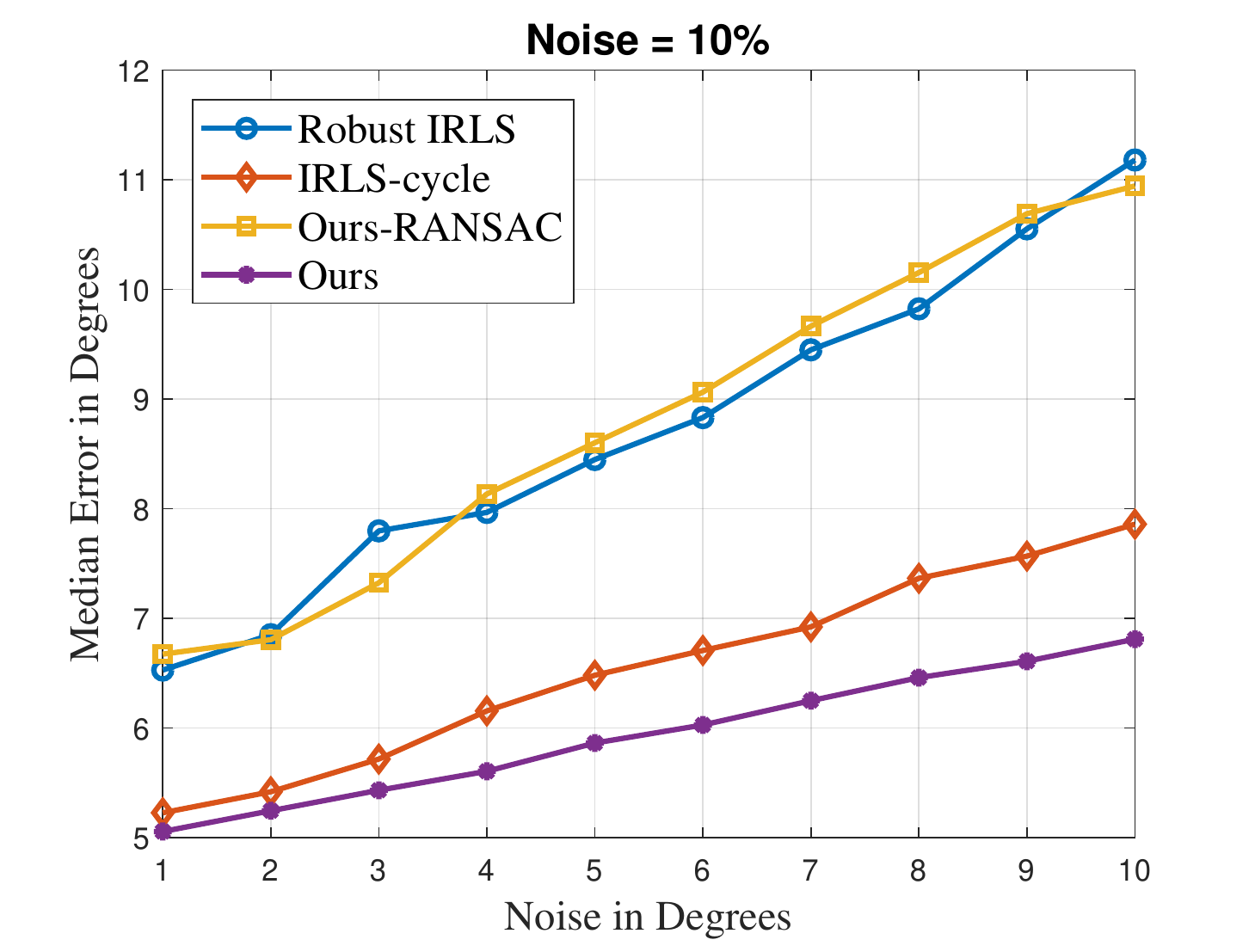}
}
\subfloat	[Noise = 20\%]	{
\includegraphics[width=.5\linewidth]{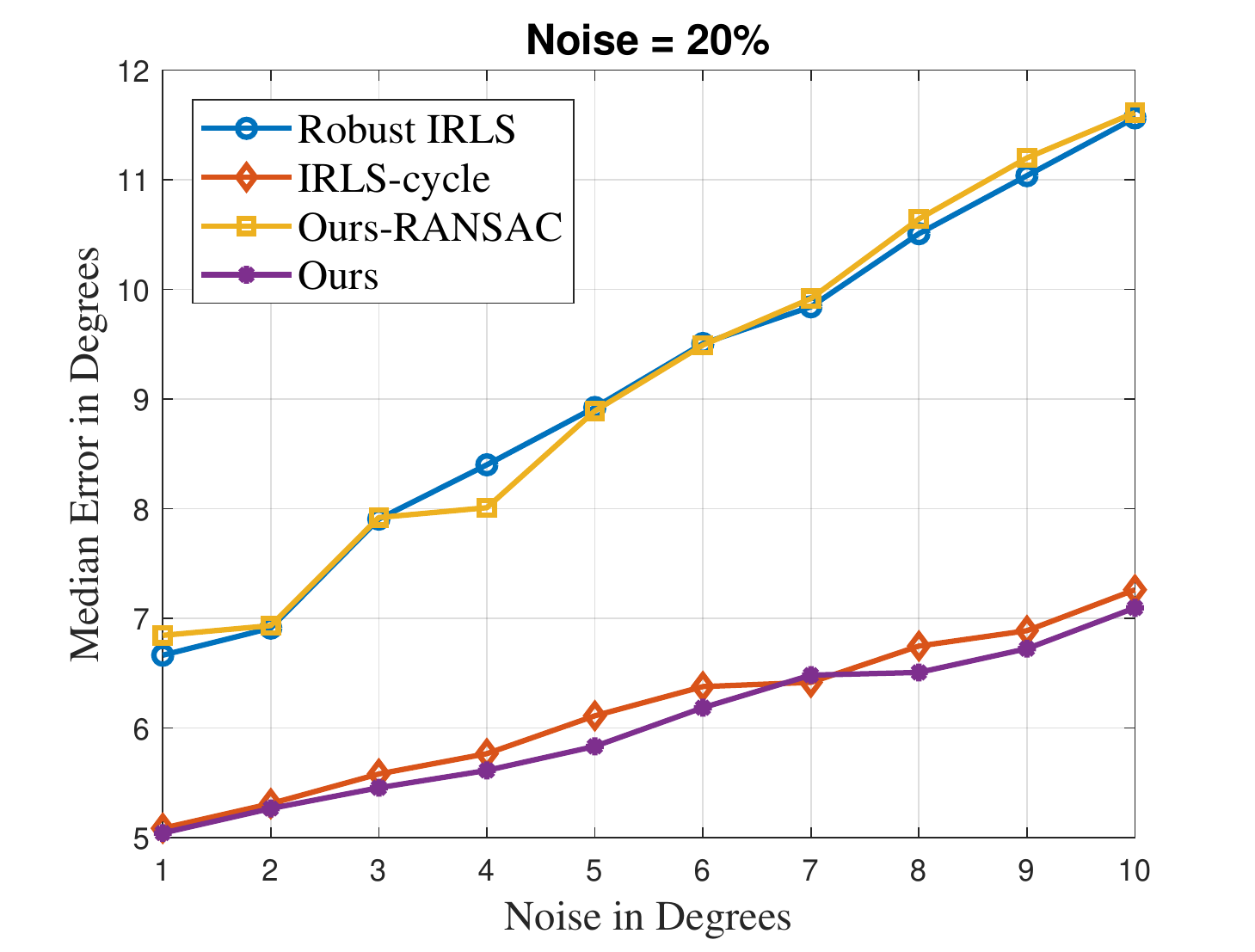}
}

\subfloat  [Noise = 30\%]    {
\includegraphics[width=0.5\linewidth]{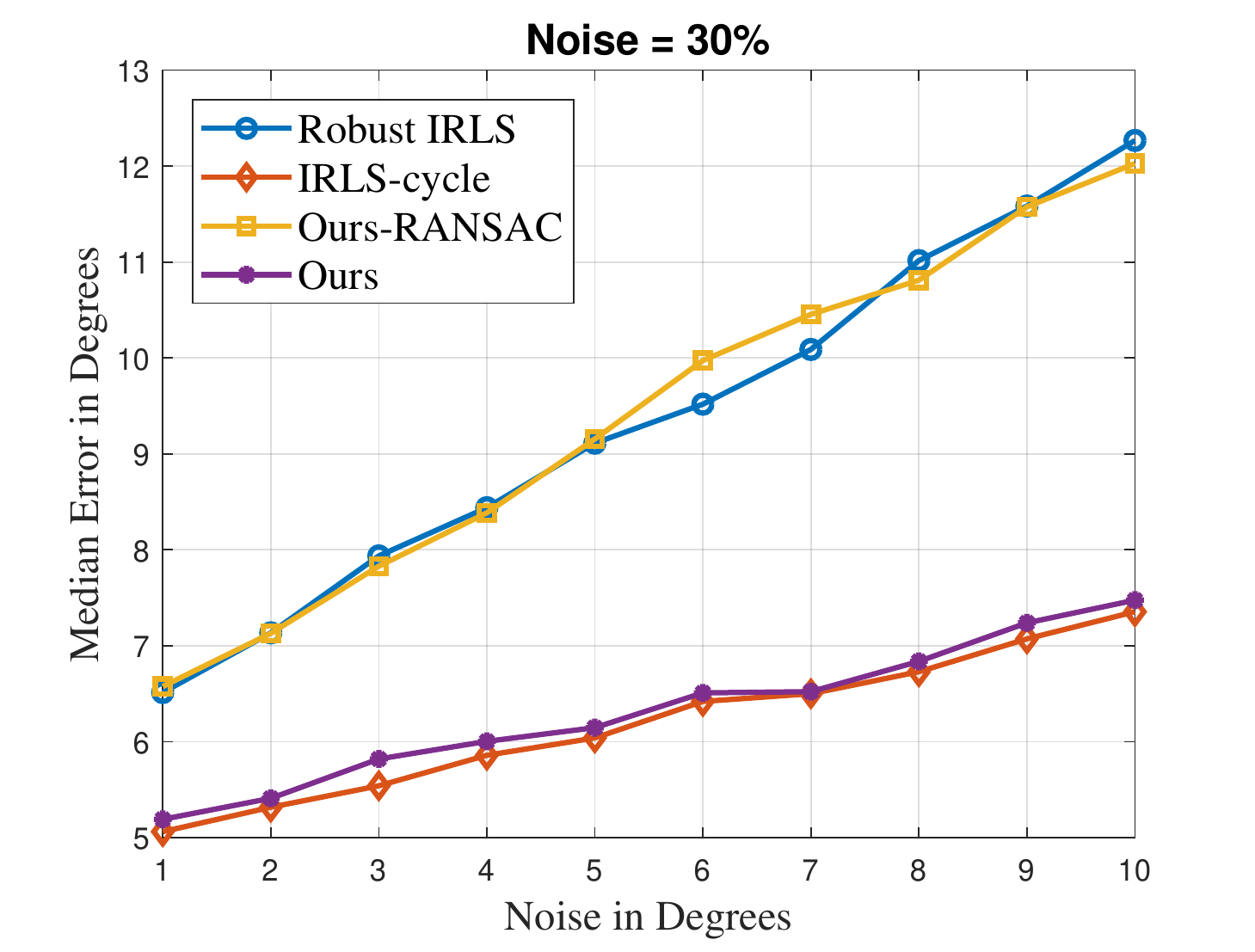}
}
\subfloat  [Noise = 40\%]    {
\includegraphics[width=0.5\linewidth]{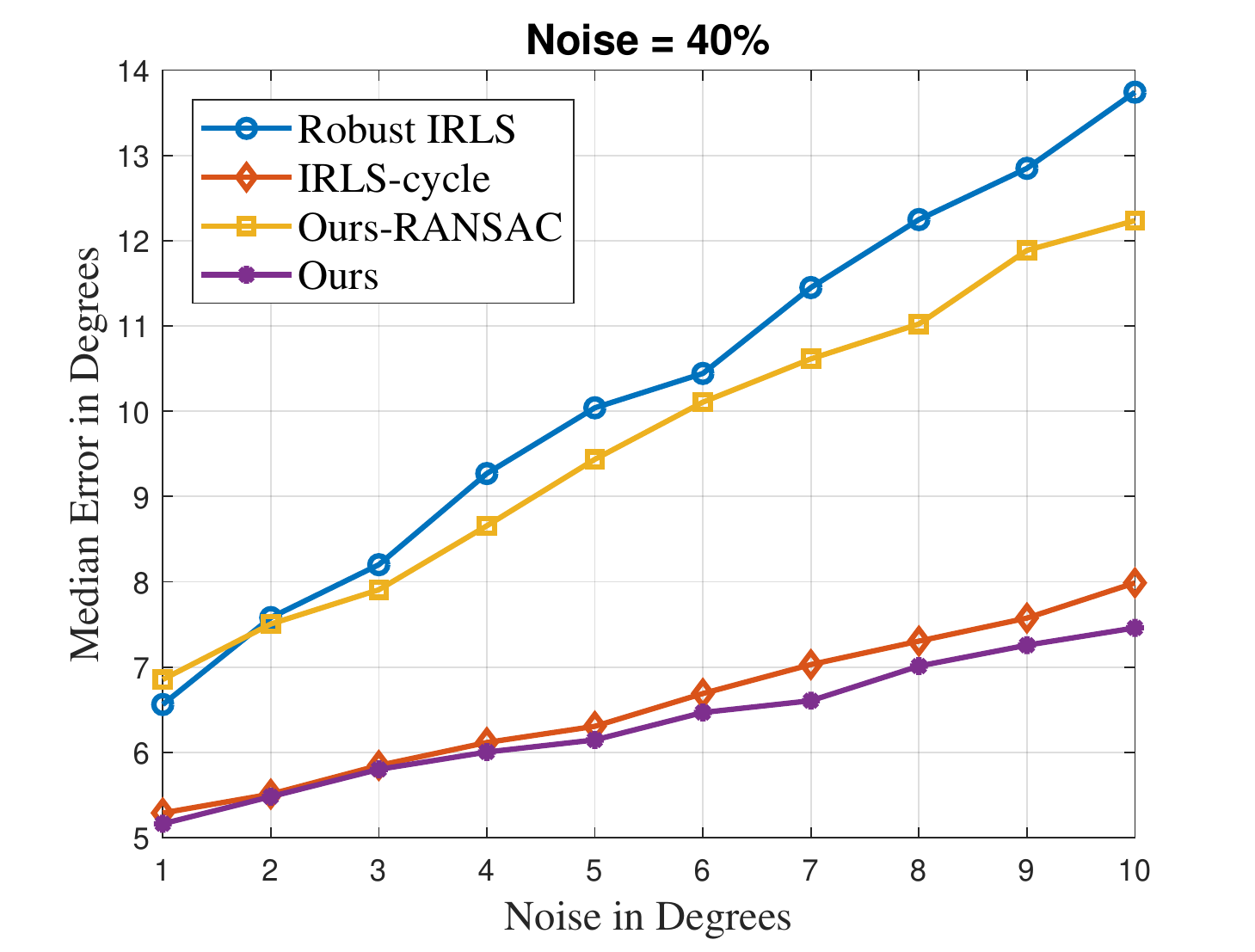}
}
    \caption{Performances with different measurement filtrations under different noise levels.}
    \label{fig:robustfig}
\end{figure}
\subsubsection{Cycle Consistency}
We also conduct experiments to analyze the erroneous measurements adjustment procedure. In the experiments, we test Robust IRLS~\cite{chatterjee2018robust} with RANSAC (original) and with our cycle consistency instead of RANSAC, Ours with RANSAC only and Ours against increasing noise. In Fig.~\ref{fig:robustfig}, we provide the performances on the optimization accuracy. It can be observed that as for the same outlier percentage, approaches with the implementation of the cycle consistency are more robust to higher level of noise. With the outlier percentage increases, the effects of RANSAC continues to decrease since there are not sufficiently many inliers for RANSAC to function as expected while cycle consistency maintains the advantage.

\clearpage
\begin{table*}[th!]
	\begin{center}
	\caption{Experiment results on the Photo Tourism Dataset~\cite{wilson_eccv2014_1dsfm}. In the table, $\Delta \bar{\text {deg}}$ and $\Delta \hat{\text {deg}}$ denote the mean and median error in degrees, respectively; runtime is in seconds and number of iterations denote the iterations to initialize + iterations for the calculation.}
	\label{table:fulltable}
	\addtolength{\parskip}{-0.2cm}
	\resizebox*{!}{0.9\textheight}{%
	\begin{tabular}{ccccccccccc}
		\toprule[0.5mm]
         & &\textbf{\makecell{IRLS\\ \cite{chatterjee2013efficient}}} & \textbf{\makecell{Robust-IRLS\\ \cite{chatterjee2018robust}}} & \textbf{\makecell{MPLS\\ \cite{lerman2019robust}}} & \textbf{\makecell{Ours-$l_2$}}&\textbf{\makecell{Ours-$l_1$}}&\textbf{\makecell{Ours-$l_{\frac{1}{2}}$}}&\textbf{\makecell{\makecell{Ours\\RANSAC}}}&
         \textbf{\makecell{Ours\\R+cycle}} &
         \textbf{\makecell{Ours\\}} \\ \midrule[0.3mm]
          \multirow{4}{*}{\rotatebox{90}{Alamo}} & $\Delta \bar{\text {deg}}$ & 3.64 & 3.67 & 3.44 & 3.54 & 3.52 & 3.44 & 3.59 & \textbf{3.36}& 3.39\\
          & $\Delta \hat{\text {deg}}$ &1.30 & 1.32 & \textbf{1.16} & 1.29 & 1.29 & 1.20 & 1.27 & \textbf{1.16}&\textbf{1.16}\\
          & runtime & \textbf{14.2} & 15.1 & 20.6 & 16.3 & 17.2 & 21.6 & 19.8 & 20.4 & 14.7\\
          & iteration &10+8 & 10+9 & 6+8 & 4+8 & 4+9 & 4+11 & 6+6 & \textbf{4+6} & \textbf{4+6}\\          
          \midrule[0.3mm]
          \multirow{4}{*}{\rotatebox{90}{Ell. Is.}} &$\Delta \bar{\text {deg}}$ & 3.04 & 2.71 & 2.61 & 2.39 &2.28 & \textbf{2.21} &2.47 &2.24 &2.25\\
          & $\Delta \hat{\text {deg}}$ &1.06 & 0.93 & 0.88 & 0.82 &0.82& 0.76& 0.90 & 0.67 & \textbf{0.64}\\
          & runtime & 3.2 & 2.8 & 4.0 &2.7& 2.8 & 3.5 & 4.1 & 2.5 &\textbf{2.3}\\
          & iteration &10+9 & 10+13 & 6+11 & 6+12 & 6+12 &6+12 &6+12 &5+10 & \textbf{5+7}\\
          \midrule[0.3mm]
          \multirow{4}{*}{\rotatebox{90}{Mont. N.D}} &$\Delta \bar{\text {deg}}$ & 1.25 & 1.22 & 1.04 & 1.12 &1.12 &1.07 & 1.18 & \textbf{0.97} & 1.04\\
          & $\Delta \hat{\text {deg}}$ &0.58 & 0.57 & 0.51 & 0.51 & 0.50 & 0.50 & 0.56 & \textbf{0.47} & \textbf{0.47}\\
          & runtime & \textbf{6.5} & 7.3 & 9.3 & 7.6 &8.1 & 8.5 & 8.6 & 7.9&7.1\\
          & iteration &10+9 & 10+13 & 6+11 & 6+12 &6+12 &6+12 &6+12 &5+8 & \textbf{5+7}\\ 
          \midrule[0.3mm]
          \multirow{4}{*}{\rotatebox{90}{Not. Da.}} &$\Delta \bar{\text {deg}}$ & 2.63 & 2.26 & 2.06 & 2.19 & 2.14& 2.03 &2.35 & \textbf{1.95} & 1.98\\
          & $\Delta \hat{\text {deg}}$ &0.78&0.71&0.67&0.71 & 0.70& 0.68 & 0.79 & 0.65 & \textbf{0.62}\\
          & runtime & \textbf{17.2}&22.5 & 31.5 & 24.5 & 25.9& 27.5 &27.6 & 23.8 &21.2\\
          & iteration &10+14 & 10+15 & 6+14 & 6+12 & 6+12& 6+12 & 6+12 & 5+12 & \textbf{5+9}\\ 
          \midrule[0.3mm]
          \multirow{4}{*}{\rotatebox{90}{Picca.}} &$\Delta \bar{\text {deg}}$ & 5.12 & 5.19 & 3.93 & 4.08 & 4.00 & 3.92 & 4.87 & \textbf{3.82} &\textbf{3.82}\\
          & $\Delta \hat{\text {deg}}$ &2.02 & 2.34 & 1.81 & 1.83 & 1.83& 1.79& 2.07 & 1.79& \textbf{1.77}\\
          & runtime & \textbf{153.5} & 170.2 & 191.9 & 180.6 & 183.9&193.5 & 200.6 & 178.7 & 173.4\\
          & iteration &10+16 & 10+19 & 6+21 & 6+19 & 6+21& 6+21 & 6+21 & \textbf{6+15} &6+16\\ 
          \midrule[0.3mm]
          \multirow{4}{*}{\rotatebox{90}{NYC Lib}} &$\Delta \bar{\text {deg}}$ & 2.71 & 2.66 & 2.63 & 2.63 & 2.60& 2.58 & 2.73 & \textbf{2.53} & 2.56\\
          & $\Delta \hat{\text {deg}}$ &1.37 & 1.30 & 1.24 & 1.20 & 1.20 & 1.17 & 1.33 & 1.19 &\textbf{1.17}\\
          & runtime & \textbf{2.5} & 2.6 & 4.5 &4.2 & 4.5& 5.1& 3.9 & 4.0 &3.7\\
          & iteration &10+14 & 10+15 & 6+14 & 6+11 & 6+12& 6+12 & 6+12 & \textbf{6+9} &\textbf{6+9}\\ 
          \midrule[0.3mm]
          \multirow{4}{*}{\rotatebox{90}{P.D.P}} &$\Delta \bar{\text {deg}}$ & 4.1 & 3.99 & 3.73 & 3.77 & 3.74& 3.69 & 3.94 &\textbf{3.60} &3.64\\
          & $\Delta \hat{\text {deg}}$ &2.07 & 2.09 & 1.93 & 1.85 & 1.84&1.85 & 1.98 & 1.83 & \textbf{1.80}\\
          & runtime & \textbf{2.8} & 3.1 & 3.5 & 3.1 & 3.4& 3.8 & 4.2 & 3.0 &2.9\\
          & iteration &10+9 & 10+13 & \textbf{6+3} & 6+7 & 6+9& 6+12 &6+12 & 5+6 &5+6\\ 
          \midrule[0.3mm]
          \multirow{4}{*}{\rotatebox{90}{Rom. For.}} &$\Delta \bar{\text {deg}}$ & 2.66 & 2.69 & 2.62 & 2.65 &2.64 &2.64 &2.71 &\textbf{2.57} & {2.60}\\
          & $\Delta \hat{\text {deg}}$ &1.58 & 1.57 & \textbf{1.37} & 1.44& 1.44 & 1.42 & 1.54 & 1.39 & {1.39}\\
          & runtime & 8.6 & 11.4 & 8.8 & 9.2 & 9.8 & 10.7 & 8.8 & \textbf{8.4} & 8.6\\
          & iteration &10+9 & 10+17 & 6+8 & 5+9 &5+12 & 5+12 & 6+12 &5+9 & \textbf{5+7}\\ 
          \midrule[0.3mm]
          \multirow{4}{*}{\rotatebox{90}{T.o.L}} &$\Delta \bar{\text {deg}}$ & 3.42 & 3.41 & 3.16 & 3.23 & 3.20& 3.14 & 3.32 & \textbf{3.03} &\textbf{3.03}\\
          & $\Delta \hat{\text {deg}}$ &2.52 & 2.50 & 2.20 & 2.12 &2.08& 2.04 & 2.20 & 2.00 & \textbf{1.98}\\
          & runtime & 2.6 & \textbf{2.4} & 2.7 & 2.5 & 2.5&2.7 & 2.5 & 2.5 &\textbf{2.4}\\
          & iteration &10+8 & 10+12& 6+7 & 6+7 & 6+10&6+12 &6+12 &5+8 &\textbf{5+6}\\ 
          \midrule[0.3mm]
          \multirow{4}{*}{\rotatebox{90}{Uni. Sq.}} &$\Delta \bar{\text {deg}}$ & 6.77 & 6.77 & 6.54 & 6.57 & 6.50& 6.44 & 6.69 & \textbf{6.20} & {6.24}\\
          & $\Delta \hat{\text {deg}}$ &3.66 & 3.85 & \textbf{3.48} & 3.70 &3.67 & 3.60 & 3.62 &3.65 &3.64\\
          & runtime & \textbf{5.0} & 5.6 & 5.7 & 5.5 & 5.8 & 6.1 & 5.8 & 5.6 & 5.4\\
          & iteration &10+32 & 10+47 & 6+21 & \textbf{6+15} &6+19 & 6+21 & 6+21 & 5+20 & {5+20}\\ 
          \midrule[0.3mm]
          \multirow{4}{*}{\rotatebox{90}{Yorkm.}} &$\Delta \bar{\text {deg}}$ & 2.6 & 2.45 & 2.47 & 2.45 & 2.44& 2.44 & 2.50 & 2.40 &\textbf{2.37}\\
          & $\Delta \hat{\text {deg}}$ &1.59 & 1.53 & 1.45 & 1.47 & 1.45 & 1.45& 1.52 & 1.36 & \textbf{1.33}\\
          & runtime & \textbf{2.4} & 3.3 & 3.9 & 3.7 &4.0 & 4.1 &3.8 &3.2& 3.1\\
          & iteration &10+7 & 10+9 & 6+7 & 5+9 & 5+11& 5+12 & 6+12 & \textbf{5+7} &\textbf{5+7}\\ 
          \midrule[0.3mm]
          \multirow{4}{*}{\rotatebox{90}{Gend. Mar.}}&$\Delta \bar{\text {deg}}$ & 39.24 & 39.41 & 44.94 & 42.76 & 39.98 & 38.70 & 40.24 & \textbf{35.94} & 36.58\\
          & $\Delta \hat{\text {deg}}$& 7.07& 7.12 & 9.87 & 9.55 & 9.26 & 9.12 & 9.93 & \textbf{7.04} & 7.21\\
          & runtime &\textbf{6.5} & 7.3 & 17.8 & 14.2 & 14.9 & 16.3 & 12.2 & 15.4 & 15.7\\
          & iteration& 10+14 & 10+19 & 6+25 & 6+16 & 6+19 & 6+25 &6+25 & \textbf{6+20} & 6+21\\
          \midrule[0.3mm]
          \multirow{4}{*}{\rotatebox{90}{Mad. Met.}}&$\Delta \bar{\text {deg}}$ & 5.3 & 4.88 & 4.65 & 4.72 & 4.62 & 4.52 & 5.43 & 4.46 & \textbf{4.38}\\
          & $\Delta \hat{\text {deg}}$ & 1.78 & 1.88 & 1.26 & 1.28 & 1.19 & 1.14& 1.64 & \textbf{1.02} & 1.07\\
          & runtime & 3.8 & \textbf{2.7} & 5.2 & 4.3 & 4.9 & 5.6 & 4.2 & 3.9 & 4.1\\
          & iteration& 10+30 & 10+12 & 6+23 & 6+15 & 6+19 & 6+25 & 6+25 & \textbf{5+21} & \textbf{5+21}\\
          \midrule[0.3mm]
          \multirow{4}{*}{\rotatebox{90}{Vien. Cat.}}&$\Delta \bar{\text {deg}}$ & 8.13 & 8.07 & 7.21 & 7.13 & 7.02 & 6.97 & 7.89 & \textbf{6.42} & 6.57\\
          & $\Delta \hat{\text {deg}}$& 1.92 & 1.76 & 2.83 & 2.32 & 2.26 & 2.29 & 2.42 & \textbf{1.55} & 1.90 \\
          & runtime & 28.3 & \textbf{13.1} & 42.6 & 26.3 & 30.2 & 35.8 & 27.4 & 28.2 & 26.3\\
          & iteration& 10+13&10+23 & 6+19 & 6+18 & 6+23 & 6+24 & 6+24 & 5+20 & \textbf{5+19}\\
          \midrule[0.3mm]
          
	\end{tabular}
	}
	\end{center}
\end{table*}
\clearpage
{
\bibliographystyle{IEEEtran}
\bibliography{dlg}
}
\end{document}